\documentclass[sigconf]{acmart}
\AtBeginDocument{%
  }

\copyrightyear{2025}
\acmYear{2025}
\setcopyright{acmlicensed}
\acmConference[CIKM '25] {Proceedings of the 34th ACM International Conference on Information and Knowledge Management}{ November 10--14, 2025}{Seoul, Republic of Korea.}
\acmBooktitle{Proceedings of the 34th ACM International Conference on Information and Knowledge Management (CIKM '25), November 10--14, 2025, Seoul, Republic of Korea}
\acmISBN{979-8-4007-2040-6/2025/11}
\acmDOI{10.1145/3746252.3761317}

\usepackage{graphicx}
\usepackage{amsmath}
\usepackage{enumitem}
\usepackage{multirow}
\usepackage{booktabs}
\usepackage{tcolorbox}
\usepackage{subcaption}
\usepackage{colortbl} 
\usepackage[normalem]{ulem}
\useunder{\uline}{\ul}{}

\definecolor{gongyaoColor}{RGB}{44,152,128}

\begin{document}

\title[Chart-CoCa: Code-driven Synthesis \& Candidate-Conditioned Generation for Chart-based QA]{Chart-CoCa: Self-Improving Chart Understanding of Vision LMs via Code-Driven Synthesis and Candidate-Conditioned Answering}

\author{Gongyao Jiang}
\affiliation{%
  \institution{The Hong Kong University of Science\\ and Technology (Guangzhou)}
  \city{Guangzhou}
  \country{China}
}
\email{jianggongyao@gmail.com}

\author{Qiong Luo}
\affiliation{%
  \institution{The Hong Kong University of Science\\ and Technology (Guangzhou)}
  \city{Guangzhou}
  \country{China}
}
\affiliation{%
  \institution{The Hong Kong University of Science\\ and Technology}
  \city{Hong Kong}
  \country{China}
}
\email{luo@cse.ust.hk}
\renewcommand{\shortauthors}{Trovato et al.}

\begin{abstract}
Vision Language Models (VLMs) often struggle with chart understanding tasks, particularly in accurate chart description and complex reasoning. 
Synthetic data generation is a promising solution, while usually facing the challenge of noise labels.
To address this challenge, we first introduce a chart synthesis pipeline that generates aligned chart-question-answer triplets through code generation and execution, ensuring the reliability of synthetic data without human intervention.
Furthermore, inspired by test-time scaling that increases inference budget and thereby improves performance, we design a candidate-conditioned answering process.
The VLM first generates multiple responses per query, and then synthesizes the final answer by contextualizing these candidates.
Experiments demonstrate significant improvements, with up to 15.50 points accuracy gain over the initial VLM, in a fully self-improving paradigm without either human-labeled data or external models.
\end{abstract}



\maketitle

\section{Introduction}
Vision Language Models (VLMs) demonstrate remarkable versatility and effectiveness across a wide range of practical applications \citep{dai2023instructblip, achiam2023gpt, dong2024internlm, Qwen2VL}. Their ability to interpret charts is particularly valuable, considering the prominence of charts in scientific literature, financial documents, and news reports \citep{kahou2017figureqa, methani2020plotqa, xu2023chartbench, liu2024mmc}. 
This task presents unique challenges that require models to describe and reason with numerical data, textual labels, and complex visual elements \citep{masry2022chartqa, wang-chartxiv-2024}.

\begin{figure}
    \centering
    \includegraphics[width=1\linewidth]{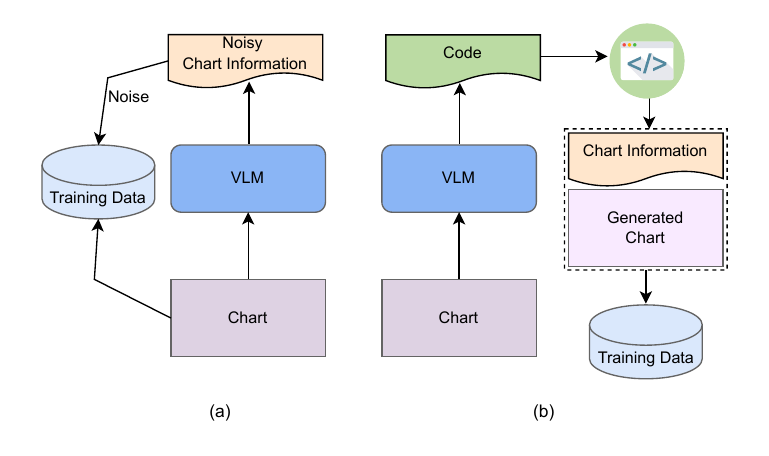}
    \caption{(a) Direct Chart-to-Text Generation. (b) Generating code first and then executing the code to obtain the pair of chart and chart information.}
    \Description{(a) Given a chart, the VLM outputs chart information with noise, which is then fed into training data for further training of the VLM. (b) Given a chart, the VLM first translates it into code. Then, through code execution, chart information and chart are obtained simultaneously to construct the training data. }
    \label{fig:intro}
\end{figure}

Enhancing VLMs' understanding capabilities largely depends on training data \citep{chen2024far}. 
However, manual chart collection and annotation are costly. 
Some studies suggest using VLMs to generate chart data for instruction tuning \citep{masry2024chartinstruct, meng2024chartassisstant}.
This kind of method enhances the performance of VLMs with more powerful external models, while the upper limit of the VLMs' own capabilities has not been explored.  
Therefore, we investigate using VLMs to autonomously generate training data and subsequently improve themselves during test time, without the need for human-labeled answers or external models.

Allowing models to generate training data by themselves confronts the challenge of data accuracy \citep{amini2025self}.
As depicted in Figure \ref{fig:intro} (a), limited by the ability of VLMs, its generated data may contain much noise, failing to provide the anticipated benefits. 
To tackle this issue, we propose employing code as the intermediary in the synthesis process, as illustrated in Figure \ref{fig:intro} (b).
This pipeline generates pairs of chart and chart information through code execution, facilitating the accurate synthesis of training data.

\begin{figure*}[t]
    \centering
    \includegraphics[width=0.86\linewidth]{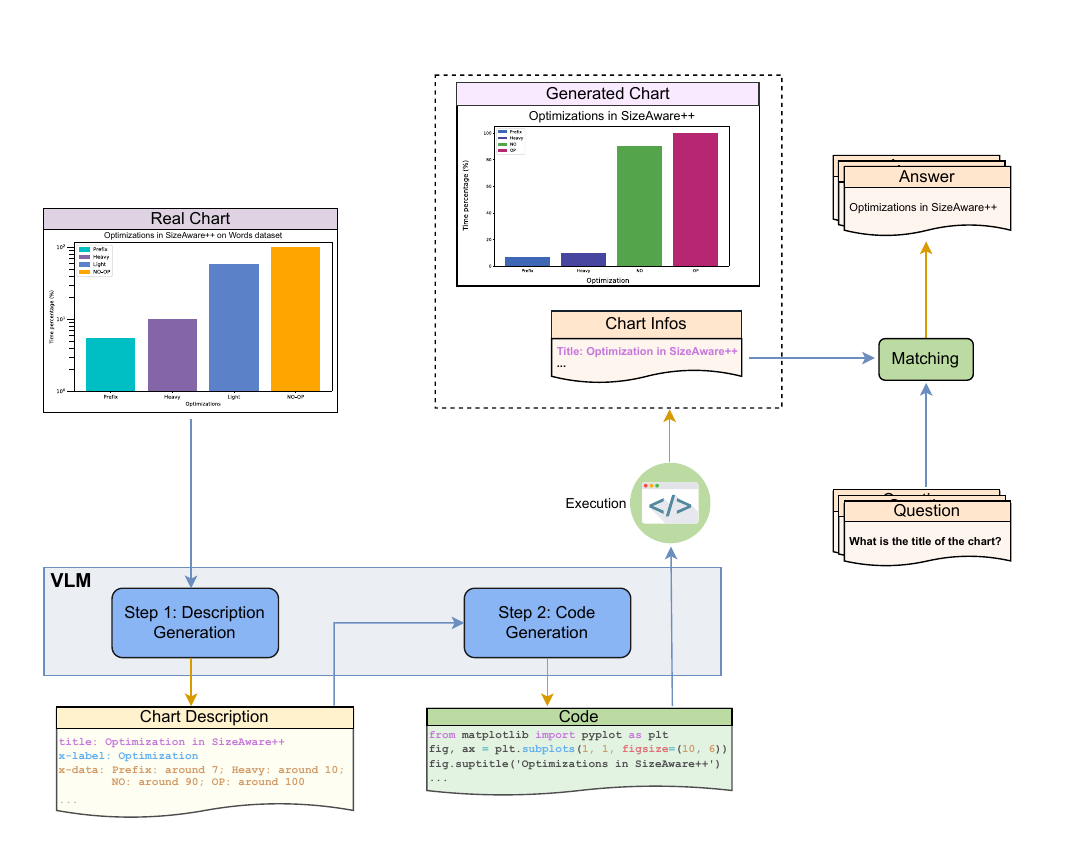}
    \caption{The pipeline of data synthesis for VLM training.}
    \Description{A real chart is first fed into a VLM to obtain its description. Then, the same VLM translates the description into Python code. The code execution obtains a generated chart and corresponding information. Answers are then generated by matching questions and the information.}
    \label{fig:overview}
\end{figure*}

As depicted in Figure \ref{fig:overview}, the VLM first describes an unlabeled chart.
The description is then passed to the VLM that generates code to create a new chart.
Then, the execution of this code results in a chart exactly matching the code. 
Meanwhile, this code facilitates the accurate extraction of chart information, such as titles, labels, and legends. 
By extracting chart information from plotting objects during code execution, we obtain information that exactly matches the chart and then acquire the pairs of question and answer that match the extracted information. 
Through code as a bridge, the process yields charts along with their corresponding pairs of question and answer.

After obtaining the generated triplets of chart-question-answer, we can directly use them to fine-tune the VLM. 
However, since the pairs of question and answer are extracted by simple rules, direct fine-tuning restricts VLM to learn to answer more complex questions. 
Inspired by the test-time scaling strategies that expand the breadth and depth of model exploration \citep{snell2024scaling, bansal2024smaller}, we propose a candidate-conditioned answering method to enhance answer accuracy.
Concretely, we allow the VLM to generate multiple responses for each chart and question. 
Then, using these generated charts, questions, and responses as input, the VLM is trained to output the correct answer. During the training phase, the answer VLM learns to discriminate between correct and incorrect responses, thereby synthesizing the correct answer. 
At test time, given a chart and a question, the initial VLM first generates multiple responses. 
Then, consistent with the training stage, given the chart, question, and multiple responses, the well-trained answer generator VLM produces the final prediction.
We name this composite method \textbf{Chart-CoCa}, as it uses \textbf{co}de as intermediary for data synthesis and adopts \textbf{ca}ndidate-conditioned generation during inference.

We evaluate our method on both descriptive and reasoning tasks for charts \citep{wang-chartxiv-2024}.
Our experiments show that our method significantly improves VLMs' chart understanding without human annotations or external models. 
Moreover, as the number of candidates increases, our method shows a continuous upward trend. 
Additionally, we compare the performance of different foundation models using our method and conduct ablation experiments to verify the effectiveness of our approach.
The code is made publicly available on \url{github.com/Zzoay/Chart-CoCa} for research purposes.

We summarize our contributions as follows:
\begin{itemize}[noitemsep, topsep=0pt] 
    \item We propose a novel chart synthesis method, utilizing code as the intermediary to ensure the accuracy of synthetic data.
    \item We present a candidate-conditioned answering method to enhance the performance of chart understanding.
    \item Our experiments demonstrate and analyze self-improvement on chart understanding tasks brought by our method.
\end{itemize}

\section{Related Work}

\noindent
\textbf{{Vision Language Models.}}
VLMs go beyond text-only LLMs, able to capture and understand the visual information present in the real world \citep{dai2023instructblip, achiam2023gpt, bai2023qwen, dong2024internlm}.
Open-source models still have a significant gap with commercial models in terms of visual understanding, especially chart understanding \citep{chen2024far, wang-chartxiv-2024}.
The closure of high-quality training data is the major reason \citep{chen2024far}.
Therefore, this study explores the chart synthesis to enhance the chart understanding of VLMs.

\noindent
\textbf{Chart-based Question Answer.}
Vision QA has attracted attention from academics in recent years.
Particularly, chart-based QA \citep{kahou2017figureqa, kafle2018dvqa, masry2022chartqa, methani2020plotqa, xia2024chartx} requires accurate perception and comprehension of charts, which poses a challenge \citep{xia2024chartx, wang-chartxiv-2024}.
Some works have demonstrated that synthetic data generation can improve VLM's chart understanding.
However, most of these works focus on using a strong model as teacher model to distill knowledge to the smaller one \citep{masry2024chartinstruct, meng2024chartassisstant}.
Our work explores a more challenging setting namely self-training or self-improving \citep{amini2025self}, requiring the VLM itself to improve its ability without external models.

\noindent
\textbf{Synthetic Data Generation.}
Data is the key to machine learning, while the manual collection of high-quality data is expensive \citep{gudivada2017data, polyzotis2018data}.
Synthetic data generation provides a promising solution to this issue, especially with LLMs' strong ability for data generation \citep{patel-etal-2024-datadreamer, long-etal-2024-llms}.
Some work has investigated how to build pipelines for data synthesis using external LLMs, bringing remarkable improvements in various tasks \citep{alpaca, wang2023self, chen2024selfplay, he2024distill}.
In contrast, we use code as an intermediary in the chart synthesis empowered by the VLM itself, ensuring accuracy without either human annotation or external models.

\noindent
\textbf{Test-time Scaling.}
Some work has demonstrated that an increase in the test-time compute can lead to a performance improvement of LLM \citep{aman-selfrefine-2023, lightman2024lets, snell2024scaling, chen2024self, bansal2024smaller, guo2025deepseek}. 
This group of studies was mainly conducted on linguistic tasks. 
Recently, some researchers have explored the self-improvement during test time of VLMs \citep{wang2024enhancing, he2024self, xu2024llavacot}. 
Inspired by these studies, we combine repeated sampling and self-checking during the inference of chart understanding.

\section{Chart-CoCa}
\label{sec:method}
We follow the categorization of chart understanding from previous work \citep{wang-chartxiv-2024}, which divides chart understanding into descriptive and reasoning tasks.
In both tasks, given the chart $\boldsymbol{c}$ and question $\boldsymbol{q}$, the VLM is tasked to predict the answer $\boldsymbol{a}$.
To be compatible with various VLMs, our method focuses on the training data generation and inference strategy rather than the model architecture.

\subsection{Code-Driven Chart Synthesis}
\noindent \textbf{Data Synthesis.} We begin with introducing a commonly-used paradigm for data augmentation, where the model directly generates answers from the inputs.
Given a chart \(\boldsymbol{c}\) and a well-defined question \(\boldsymbol{q}\), such as ``What is the title of the plot?'', the VLM \(\mathcal{M}\) predicts answer \(\boldsymbol{\hat{a}}\). 
However, the prediction \(\boldsymbol{\hat{a}}\) might be incorrect, thereby introducing noise into model training.

\noindent\textbf{Description Generation.}
As depicted in Figure \ref{fig:overview}, we initially task the VLM \(\mathcal{M}\) with generating a detailed description \(\boldsymbol{d}\) of the given chart \(\boldsymbol{c}\). 
While this process may also introduce noise, such noise is acceptable, as our goal is not to accurately reconstruct the chart. 
Instead, we prompt the VLM to include comprehensive information of the chart, regardless of the accuracy, to make the subsequent code generation be as detailed as possible.
This process can be formulated as:
\begin{equation}
    \boldsymbol{d} = \mathcal{M} \left( \boldsymbol{c} \right)
    \label{eq:chart2text}
\end{equation}

\noindent\textbf{Code Generation.}
After obtaining the description of the chart, we transform this text-based information into executable code.
In this step, we input the generated description \(\boldsymbol{d}\) into the language model component \(\mathcal{M}_{\text{LM}}\) of the VLM \(\mathcal{M}\) to produce code that aligns with the description and can generate a new chart. 
The code is constrained to use the ``Python Matplotlib'' library for plotting, in order to facilitate easy execution and information extraction.
This process is defined by Equation \ref{eq:text2code}.
\begin{equation}
    \boldsymbol{p} = \mathcal{M}_{LM} \left( \boldsymbol{d} \right)
    \label{eq:text2code}
\end{equation}

\noindent\textbf{Code Execution.}
After obtaining the generated code, we execute it to produce a chart. 
However, limited by the ability of \(\mathcal{M}_{\text{LM}}\), the code execution may result in errors. 
To mitigate this problem, we simply set a maximum number of attempts and simply retry the code generation (Equation \ref{eq:text2code}) and code execution.
If the code executes successfully or the maximum number of attempts is reached, we proceed to the next instance.

Each plotting object in Python's Matplotlib library contains detailed information. 
For instance, a line graph object includes attributes such as titles, x-axis and y-axis labels, and lines. 
We directly extract these elements during code execution. 
This chart generation and information extraction is defined as:
\begin{equation}
    \boldsymbol{c}^{*},\boldsymbol{e} = \text{Execution} \left( \boldsymbol{p} \right)
    \label{eq:code2chart}
\end{equation}

\begin{table*}[t]
    \small
    \setlength{\tabcolsep}{1pt}
    \begin{tabular}{lll}
        \toprule
        \textbf{Element} &  \textbf{Function} &  \textbf{Example Question} \\
        \midrule
        Layout & fig, axes = plt.subplots(rows, columns) & What is the subplot layout of this chart? \\
                       & & What is the number of subplots? \rule{0pt}{7pt} \\ \addlinespace[5pt] 
        Title & axes[i,j].get\_title() & What is the title of the subplot at row i and column j? \\ \addlinespace[5pt]
        X label & axes[i,j].get\_xlabel() & What is the label of x-axis of the subplot at row i and column j? \\ \addlinespace[5pt] 
        Y label & axes[i,j].get\_ylabel() & What is the label of y-axis of the subplot at row i and column j? \\ \addlinespace[5pt] 
        Legend & axes[i,j].get\_legend() & What are the names of the labels in the legend of the subplot? \\
               & & How many discrete labels are there in the legend of the subplot at row i and column j? \rule{0pt}{7pt} \\ \addlinespace[5pt] 
        X ticks & axes[i,j].get\_xticklabels() & What is the leftmost labeled tick on the x-axis of the subplot at row i and column j? \\
                      & & What is difference between consecutive numerical tick values on the x-axis of the chart...? \rule{0pt}{7pt} \\ \addlinespace[5pt] 
        Y ticks & axes[i,j].get\_yticklabels() & What is the lowest labeled tick on the y-axis of the subplot at row i and column j? \\
                      & &  What is difference between consecutive numerical tick values on the y-axis of the chart...? \rule{0pt}{7pt} \\ \addlinespace[5pt] 
        Colorbar & axes[i,j].get\_images()[0].colorbar & What is the maximum value of the tick labels on the colorbar of the subplot at row i...? \\
                 & & What is the difference between the maximum and minimum values of the colorbar of ...? \rule{0pt}{7pt} \\ \addlinespace[5pt] 
        Lines & axes[i,j].lines & How many lines are in the subplot at row i and column j? \\
        \bottomrule
    \end{tabular}
    \caption{Elements that can be directly extracted via the code, corresponding functions, and sample questions.}
    \label{tab:quest}
\end{table*}

\noindent\textbf{QA Generation.}
We use the descriptive questions in CharXiv \citep{wang-chartxiv-2024} as seeds for question enrichment and answer extraction, as these questions only involve basic elements such as the layout, title, and labels of the chart.
This nature meets the requirements of seeds for data synthesis: simplicity, virtually no design cost, easy accessibility to answers, and suitability for the majority of charts.
We get the information related to each question through reflection at code execution.
Table \ref{tab:quest} shows example questions, their associated chart elements and reflection functions.
This information can directly and precisely answer the descriptive questions.
For example, when asked the layout of the chart, the attribute of the item ``layout'' directly answers the question.
By this manner, we obtain accurate answers $\boldsymbol{a}^{*}$ by matching questions $\boldsymbol{q}^{*}$ with these extracted information $\boldsymbol{e}$, as formulated in Equation \ref{eq:code2answer}.


\begin{equation}
    \boldsymbol{a}^{*} = \text{Match} \left( \boldsymbol{q}^{*}, \boldsymbol{e} \right)
    \label{eq:code2answer}
\end{equation}

To enrich the diversity of training data, as the rephrasing ability of LLMs has been demonstrated in previous work \citep{bai2023qwen, deng2024rephraserespondletlarge, zhang-etal-2024-llm-assisted}, we further use the VLM itself to rephrase the questions and answers, forcing it to retain the original meaning through prompts.

Since the answers are extracted directly from the executed code, and the corresponding chart is generated directly from the same code, we obtain the triplets of chart-question-answer of high accuracy.
A direct approach is to use the triplets $\left< \boldsymbol{c}^{*}, \boldsymbol{q}^{*}, \boldsymbol{a}^{*} \right>$ for fine-tuning a VLM, which takes chart and question as input and expects the VLM to output the answer.
However, as the question is simply defined and relatively fixed, the VLM may not learn to address flexible questions. 
Fortunately, it has been demonstrated that with the increase of inference budget \citep{snell2024scaling, bansal2024smaller}, the ability to solve diverse problems improves. 
Inspired by that, we design a test-time scaling strategy to enhance the VLM's ability to understand charts.

\subsection{Candidate Generation}
Previous studies demonstrated that by simply repeatedly sampling responses to a single query, the coverage of correct answers increases \citep{snell2024scaling, chen2024self}.
Therefore, we first have the VLM generate multiple candidates $\left< \boldsymbol{r}^{*}_1, \boldsymbol{r}^{*}_2,..., \boldsymbol{r}^{*}_k \right>$ to the given generated chart $\boldsymbol{c}^{*}$ and question $\boldsymbol{q}$,
by repeatedly sampling tokens from the probability distribution output by a single forward pass.

\begin{equation}
    \left< \boldsymbol{r}^{*}_1, \boldsymbol{r}^{*}_2, \dots, \boldsymbol{r}^{*}_k \right> = \mathcal{M} \left( \boldsymbol{c}^{*}, \boldsymbol{q}^{*} \right)
    \label{eq:candidate}
\end{equation}
To facilitate presentation, $\left< \boldsymbol{r}^{*}_1, \boldsymbol{r}^{*}_2, \dots, \boldsymbol{r}^{*}_k \right>$ are denoted as $\boldsymbol{r}^{*}_{1,\dots,k}$ in the subsequent sections.

\subsection{Candidate-Conditioned Answering}
Next, we refine the final answer based on the candidates. 
A direct method that utilizes the candidates is majority voting, which has demonstrated a strong baseline \citep{lewkowycz2022solving, wangself2023}.
However, the consensus is not necessarily indicative of correctness, demanding a more effective way to identify the correct answer \citep{snell2024scaling}. 
Best-of-N methods typically employ a reward model to score candidates and select the one with the highest score as the final prediction, showing remarkable performance \citep{lightman2024lets}.
However, this approach is constrained by the coverage of possible answers by the candidates.
Therefore, we introduce greater flexibility in answer modeling by enabling a VLM to generate the final answer conditioned on the candidates.

\begin{figure}[t]
    \centering
    \includegraphics[width=0.88\linewidth]{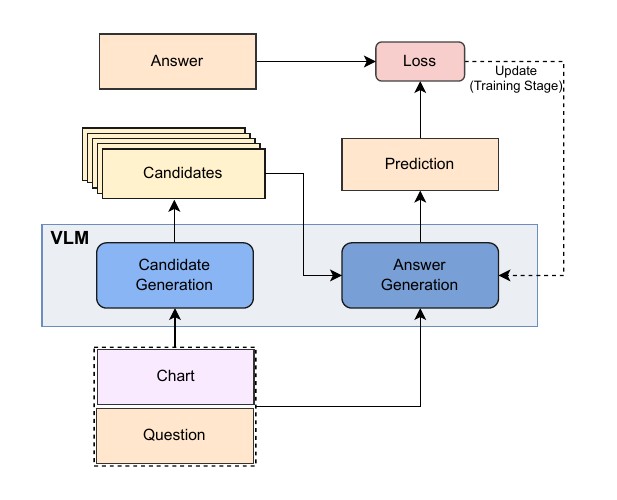}
    \caption{The process of candidate-conditioned generation. The initial VLM first generates answer candidates. Then, the fine-tuning (training-stage) / fine-tuned (test-time) VLM generates the final prediction by using these candidates as references.}
    \Description{Given a pair of (generated) chart and question, a VLM first generates answer candidates. Then, given the triplet of these three, a fine-tuning VLM in the training stage or a fine-tuned VLM after training generates the final answer.}
    \label{fig:infer}
\end{figure}

\noindent\textbf{Training.}
To enable the answer model to handle the integrated input and output the expected answer, we fine-tune the initial VLM by using our synthetic data.
As depicted in Figure \ref{fig:infer}, given the generated chart $\boldsymbol{c}^{*}$, question$\boldsymbol{q}^{*}$, and candidates $\boldsymbol{r}^{*}_{1,\dots,k}$, we have the same VLM $\mathcal{M}$ output the final answer $\boldsymbol{a}^{*}$:

\begin{equation}
    \boldsymbol{\hat{a}^{*}} = \mathcal{M} \left( \boldsymbol{c}^{*}, \boldsymbol{q}^{*}, \boldsymbol{r}^{*}_{1,\dots,k} \right)
    \label{eq:answer}
\end{equation}
The prediction $\boldsymbol{\hat{a}^{*}}$ is then compared with ground truth $\boldsymbol{a}^{*}$ to calculate the loss. 
Based on the computed loss, we then update the parameters of the model $\mathcal{M}$ iteratively and obtain the answer model $\mathcal{M}^\text{ANS}$ at the end of training.

\noindent\textbf{Inference.}
Depicted in Figure \ref{fig:infer}, the inference process follows the same flow as the training stage. 
Given the real chart $\boldsymbol{c}$ and question $\boldsymbol{q}$, the initial VLM $\mathcal{M}$ generates $k$ responses.
Then, the trained answer VLM $\mathcal{M}^\text{ANS}$ produces the final answer $\boldsymbol{\hat{a}}$, as formulated in Equation \ref{eq:infer}.

\begin{equation}
\begin{aligned}
    \boldsymbol{r}_{1,\dots,k} &= \mathcal{M} \left( \boldsymbol{c}, \boldsymbol{q} \right)\\
    \boldsymbol{\hat{a}} =  &\mathcal{M}^{\text{ANS}} \left( \boldsymbol{c}, \boldsymbol{q}, \boldsymbol{r}_{1,\dots,k} \right)
\end{aligned}
\label{eq:infer}
\end{equation}
\section{Experiment}

\begin{table*}[t]
\centering
\small
\setlength{\tabcolsep}{8pt}
\begin{tabular}{lrrrrrrrrrrr}
\toprule
\multicolumn{1}{c}{}                        & \multicolumn{6}{c}{\textbf{Descriptive Questions}}                                                                                                                                                                                                                                                     & \multicolumn{5}{c}{\textbf{Reasoning Quesitons}}                                                                                                                                                                                                                                                                                                                                         \\ \cmidrule(lr){2-7} \cmidrule(lr){8-12} 
\multicolumn{1}{c}{\multirow{-2}{*}{\textbf{Method}}} & \multicolumn{1}{c}{\begin{tabular}[c]{@{}c@{}}\textbf{Info.} \\ \textbf{ Extr.} \end{tabular}} & \multicolumn{1}{c}{\textbf{Enum.}} & \multicolumn{1}{c}{\begin{tabular}[c]{@{}c@{}}\textbf{Patt.}  \\ \textbf{Rec.}\end{tabular}} & \multicolumn{1}{c}{\textbf{Cntg.}} & \multicolumn{1}{c}{\textbf{Comp.}} & \multicolumn{1}{c}{\cellcolor[HTML]{EFEFEF}\textbf{All}} & \multicolumn{1}{c}{\begin{tabular}[c]{@{}c@{}}\textbf{Text in} \\ \textbf{Chart}\end{tabular}} & \multicolumn{1}{c}{\begin{tabular}[c]{@{}c@{}}\textbf{Text in} \\ \textbf{General}\end{tabular}} & \multicolumn{1}{c}{\begin{tabular}[c]{@{}c@{}}\textbf{Num. in} \\ \textbf{Chart} \end{tabular}} & \multicolumn{1}{c}{\begin{tabular}[c]{@{}c@{}}\textbf{Num. in} \\ \textbf{General}\end{tabular}} & \multicolumn{1}{c}{\cellcolor[HTML]{EFEFEF}\textbf{All}} \\ 
\midrule
Initial Model          & 69.40          & 40.52          & 44.76          & 61.83          & 19.64          & \cellcolor[HTML]{F2F2F2}54.10                & 21.82                & 40.40                & 28.45                & 14.85                & \cellcolor[HTML]{F2F2F2}23.60                \\
CoT               & 65.54          & 46.57          & 42.58          & 58.02          & 10.71          & \cellcolor[HTML]{F2F2F2}53.23                & 25.68                & 39.39                & 29.74                & 20.52                & \cellcolor[HTML]{F2F2F2}26.80                \\
Majority Voting   & 70.23          & 57.30          & {\ul 55.02}    & 65.14          & {\ul 23.66}    & \cellcolor[HTML]{F2F2F2}61.38                & 26.36                & 41.41                & 28.02                & 19.65                & \cellcolor[HTML]{F2F2F2}26.70                \\
Self-Verification & 68.32          & 49.64          & 53.71          & 62.60          & 13.84          & \cellcolor[HTML]{F2F2F2}57.25                & {\ul 29.09}          & {\ul 41.41}          & {\ul 30.17}          & 19.21                & \cellcolor[HTML]{F2F2F2}{\ul 28.30}          \\[1pt]
\hline \addlinespace[0.5pt] 
\multicolumn{12}{c}{Code-mediated Chart Synthesis}                                                                                                                                                                                                                                                                   \\
\hline \addlinespace[1pt] 
Directly FT         & 67.73          & 61.02          & 46.29          & 44.53          & 17.86          & \cellcolor[HTML]{F2F2F2}58.95                & 19.09                & 35.35                & 27.16                & 16.16                & \cellcolor[HTML]{F2F2F2}21.90                \\
Reward Model         & {\ul 71.95}    & {\ul 68.60}    & 54.37          & {\ul 65.65}    & 21.43          & \cellcolor[HTML]{F2F2F2}{\ul 65.45}          & 27.73                & 40.40                & 29.31                & {\ul 22.27}          & \cellcolor[HTML]{F2F2F2}28.10                \\
Chart-CoCa  & \textbf{74.20} & \textbf{75.46} & \textbf{58.95} & \textbf{68.19} & \textbf{26.79} & \cellcolor[HTML]{F2F2F2}\textbf{69.60}       & \textbf{31.59}       & \textbf{44.44}       & \textbf{32.33}       & \textbf{25.33}       & \cellcolor[HTML]{F2F2F2}\textbf{31.60}    \\
\bottomrule
\end{tabular}
\caption{Evaluation results. There are two types of chart understanding tasks. Descriptive tasks contains ``Information Extraction,'' ``Enumeration,'' ``Pattern Recognition,'' ``Counting,'' ``Compositionality.'' Reasoning tasks includes ``Text in Chart,'' ``Text in General,'' ``Number in Chart,'' ``Number in General.''}
\label{tab:res}
\end{table*}

\subsection{Setup}
\noindent \textbf{Dataset.} We evaluate on the CharXiv dataset \citep{wang-chartxiv-2024}, which contains two task categories (descriptive and reasoning) designed to comprehensively assess chart understanding capabilities of VLMs. 
Compared to prior datasets \citep{masry2022chartqa, xu2023chartbench, xia2024chartx}, CharXiv demonstrates superior data diversity and annotation quality. 
Moreover, compared with the multiple-choice questions, CharXiv's open-ended questions are less likely to have their answers hit through random trial, making them more suitable for studying test-time scaling.
The dataset is officially divided into two splits: a validation set with 1,000 charts and a test set with 1,323 charts. 
In the validation set of CharXiv, each chart contains 5 sets of question-answer pairs (4 descriptive and 1 reasoning), while the answers in the test have not been publicly released by CharXiv.
Therefore, to ensure evaluation integrity and facilitate the follow-up work to ours, we use the validation set of CharXiv for model evaluation and utilize the charts from CharXiv's test set as seeds for synthetic data generation.
Additional dataset statistics and examples are provided in Appendix \ref{sec:data}.

\noindent \textbf{Evaluation.} 
We follow the CharXiv benchmarking protocol for evaluation. 
This includes maintaining an open-ended response format without structural constraints along with automated answer verification. 
Following the protocol, GPT-4o performs pairwise comparison between model predictions and ground truth answers to determine correctness. 
\citet{wang-chartxiv-2024} have discussed the reliability and potential biases of GPT-4o for the CharXiv's evaluation, and they found that GPT-4o can complete evaluation tasks well, has strong consistency between different trials, and rarely makes mistakes.
The final accuracy metric is calculated as the percentage of correct answers.

\noindent \textbf{Implementation Details.}
We fine-tuned VLMs over 3 epochs using 2 $\times$ A100 80G GPUs, each with a batch size of 1. 
To increase the effective batch size for the one-step parameter optimization, we set the gradient accumulation steps to 8. 
The learning rate was established at 2e-5 with a warmup ratio of 0.1. 
For other hyperparameters, we adhered to the default settings of each VLM. 
To mitigate the GPU memory consumption, we utilized ZeRO stage 3 \citep{rajbhandari2020zero} and allow offloading some memory to the CPU. 
We downloaded and loaded the model checkpoints through the Huggingface platform \citep{wolf2019huggingface}.
We used the default temperature setting and empirically set top\_p to 0.6, ensuring the stability of the LLMs' output while retaining diversity. 
To achieve rapid inference and memory efficiency, we employed the vLLM and LMDeploy libraries to deploy our API services.
For models with 7 to 8 billion parameters, the complete training data generation process requires approximately 3 hours, while the fine-tuning process takes around 2 hours.
For the 26B model, we used about 6 hours to generate training data and employed LoRA (rank=16) for parameter-efficiency fine-tuning, which takes approximately 5 hours.
In the inference stage, the 7B models took an average of 1.82 seconds to complete the final generation of an answer, while the 26B model took an average of 4.35 seconds.

\subsection{Comparison With Baselines}

We establish baseline comparisons by implementing multiple self-improvement strategies on the InternVL2-8B foundation model \citep{chen2024far}. 
Approaches without synthetic data includes: (a) Chain-of-Thought (CoT) prompting \citep{wei2022chain}, (b) Majority Voting with 5 candidates, and (c) Self-Verification \citep{weng2023large} with 5 candidates. 
For synthetic approaches leveraging code-mediated chart synthesis in Section\ref{sec:method}: (d) Direct Fine-tuning (FT) on synthetic data, (e) Reward Modeling for the selection from 5 candidates \citep{cobbe2021training, lightman2024lets}, and (f) Our proposed answer generation conditioned on 5 candidates (Chart-CoCa).

Table \ref{tab:res} presents the evaluation results. The initial model achieves moderate performance on both descriptive (54.10) and reasoning tasks (23.60), highlighting the inherent limitations of VLMs in chart understanding. 
CoT shows a slight improvement in reasoning tasks (26.80 vs 23.60) but at the cost of degraded performance on descriptive tasks (53.23 vs 54.10). 
Majority voting, which aggregates five candidate responses, significantly boosts descriptive task performance (61.38 vs 54.10) but offers only marginal gains in reasoning tasks (26.70 vs 23.60). 
This suggests that statistical aggregation is effective for factual tasks but insufficient for complex reasoning. 
Self-verification, while leveraging the same five candidates, underperforms majority voting on descriptive tasks (57.25 vs 61.38) and achieves comparable reasoning performance (28.30 vs 26.70). 
This indicates that while self-verification mechanisms can moderately enhance reasoning, their effectiveness is limited by the model's inherent capabilities, particularly in descriptive tasks where majority voting proves more robust.

Among methods utilizing synthetic data, directly fine-tuning achieves notable gains in descriptive tasks (58.95 vs 54.10) but suffers a drop in reasoning performance (21.90 vs 23.60), demonstrating that naive fine-tuning may impair the model's original reasoning abilities. 
The reward modeling for the candidate selection partially mitigates this issue (65.45 descriptive, 28.10 reasoning) by selecting better answers from candidates, but its performance remains limited by the quality of the initial candidates. 
Our Chart-CoCa method achieves state-of-the-art performance on both descriptive (69.60) and reasoning tasks (31.60), surpassing all baselines by significant margins ($\frac{69.60-54.10}{54.10}$=28.65\% and $\frac{31.60-23.60}{23.60}$=33.90\% over the initial model, respectively). 
This demonstrates the effectiveness of our method in simultaneously enhancing both descriptive and reasoning capabilities.

Among all sub-tasks, our method achieves the most substantial gains in enumeration (+34.94) and pattern recognition (+14.19) over the initial model, suggesting its particular strength in helping VLMs systematically identify discrete chart elements and discern underlying data trends. 
While information extraction shows more modest gains (+4.80), this aligns with the inherent challenge of precise visual grounding in pre-trained VLMs. 
The persistent difficulty in compositionality tasks (26.79) highlights fundamental limitations in processing numerical scales and symbolic chart components.
Furthermore, reasoning tasks exhibit lower absolute performance (31.60) compared to descriptive tasks (69.60).
Nevertheless, our Chart-CoCa mechanism delivers consistent improvements on all sub-tasks, showing the effectiveness of code-based synthesis and candidate-based answering.

\subsection{Impact of Candidate Number}
\begin{figure}
    \centering
    \includegraphics[width=1\linewidth]{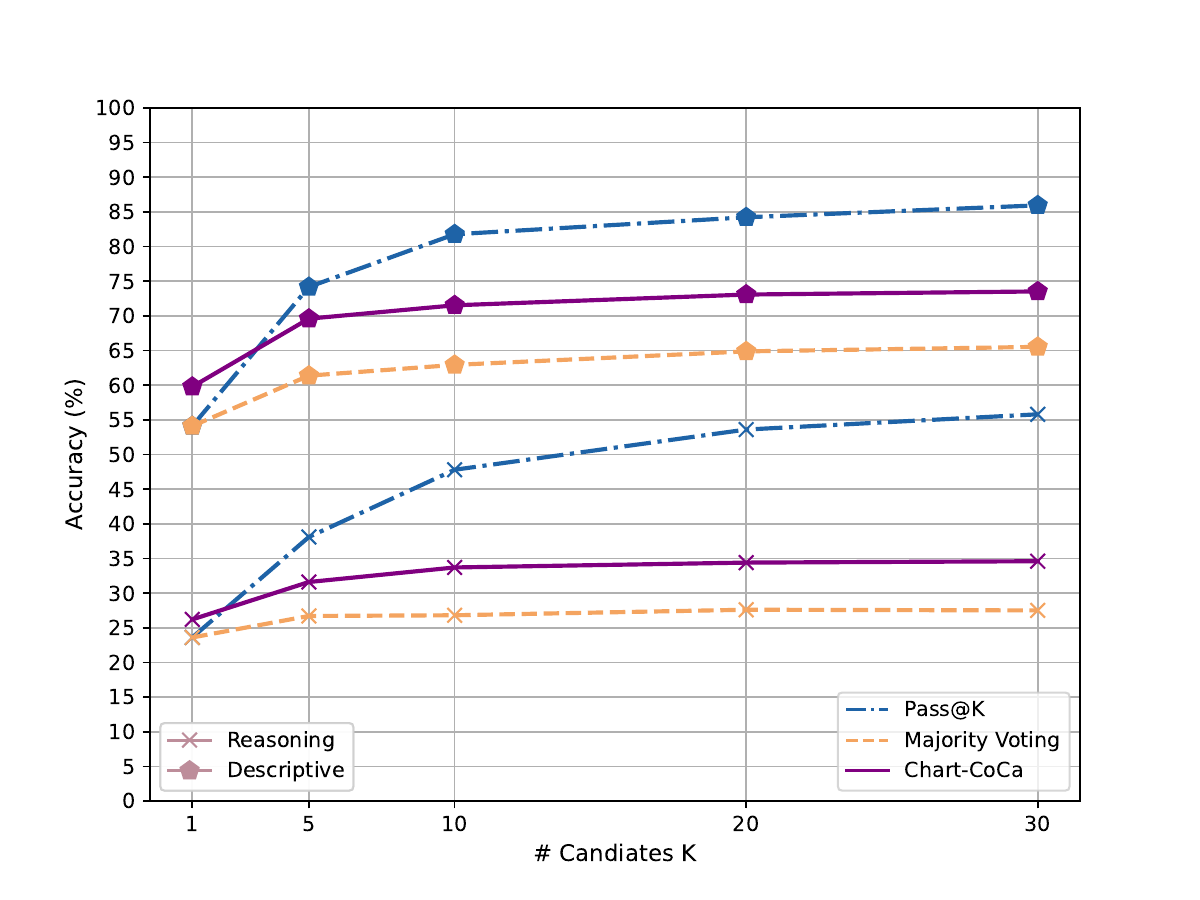}
    \caption{Performance of Pass@K, Majority Voting, and our Chart-CoCa for chart understanding tasks across increasing candidate counts K. Results are shown for both descriptive and reasoning tasks.}
    \label{fig:scaling}
\end{figure}
We investigate the test-time scaling law on chart understanding tasks.
Specifically, we task a VLM with generating multiple responses (K candidates) for a given chart-question pair. 
These candidates are evaluated under three paradigms: 1) Pass@K, measuring whether any candidate contains the correct answer; 2) majority voting accuracy; and 3) accuracy of our Chart-CoCa. 
The scaling trends across K from 1 to 30 are shown in Figure \ref{fig:scaling}.

The coverage of correct answers (Pass@K) exhibits strong scaling with K. 
For descriptive tasks, Pass@K rises from 54.1\% at K=1 to 85.95\% at K=30, demonstrating VLMs’ inherent capacity to generate correct answers given sufficient trials. 
For reasoning tasks, Pass@K also rises significantly, improving from from 23.6\% (K=1) to 55.8\%.  
Although there is still a large room on reasoning tasks, the significant increase in coverage demonstrates the prospects of test-time scaling. 

Though majority voting improves accuracy, its gains plateau early and remain far below Pass@K ceilings. 
For descriptive tasks, majority accuracy plateaus at 64.88\% (K=20) versus 85\% Pass@K. Reasoning tasks show even starker limitations: majority voting peaks at 26.80\% (K=10) versus 47.80\% Pass@K. 
Notably, majority accuracy stabilizes at K$\geq$10 for both tasks (descriptive: $\leq$2.65 points gain from K=10 to 30; reasoning: $\leq$0.9 points), implying diminishing returns from additional candidates under this strategy. 
This highlights the inefficiency of naive consensus mechanisms in exploiting the full potential of multiple candidates.

Our Chart-CoCa demonstrates significant improvements, highlighting its effectiveness in test-time scaling. 
Interestingly, even at K=1, Chart-CoCa achieves a performance boost over baselines, improving descriptive tasks from 54.1\% to 59.78\% and reasoning tasks from 23.6\% to 26.2\%. 
This suggests inherent self-verification and self-correction capabilities. 
As the candidate number K increases, the gains become more pronounced, particularly at K$\geq$5, where Chart-CoCa outperforms majority voting by a clear margin, reaching 69.6\% compared to 61.38\% for descriptive tasks at K=5. 
While our method converges at K$\geq$20, it does more gradually compared to majority voting, which plateaus earlier. 
Given the trade-offs between computational costs, latency, and performance, K=5 or K=10 emerges as a practical choice for real-world applications. 
Although Chart-CoCa still falls short of the ideal Pass@K ceiling, it significantly enhances VLM performance. 
This underscores the potential of candidate-based inference and leaves room for further exploration in test-time scaling strategies.  

\begin{table*}[t]
\centering
\small
\setlength{\tabcolsep}{8pt}
\begin{tabular}{lrrrrrrrrrrr}
\toprule
\multicolumn{1}{c}{}                        & \multicolumn{6}{c}{\textbf{Descriptive Questions}}                                                                                                                                                                                                                                                     & \multicolumn{5}{c}{\textbf{Reasoning Questions}}                                                                                                                                                                                                                                                                                                                                         \\ \cmidrule(lr){2-7} \cmidrule(lr){8-12} 
\multicolumn{1}{c}{\multirow{-2}{*}{\textbf{Method}}} & \multicolumn{1}{c}{\begin{tabular}[c]{@{}c@{}}\textbf{Info.} \\ \textbf{ Extr.} \end{tabular}} & \multicolumn{1}{c}{\textbf{Enum.}} & \multicolumn{1}{c}{\begin{tabular}[c]{@{}c@{}}\textbf{Patt.}  \\ \textbf{Rec.}\end{tabular}} & \multicolumn{1}{c}{\textbf{Cntg.}} & \multicolumn{1}{c}{\textbf{Comp.}} & \multicolumn{1}{c}{\cellcolor[HTML]{EFEFEF}\textbf{All}} & \multicolumn{1}{c}{\begin{tabular}[c]{@{}c@{}}\textbf{Text in} \\ \textbf{Chart}\end{tabular}} & \multicolumn{1}{c}{\begin{tabular}[c]{@{}c@{}}\textbf{Text in} \\ \textbf{General}\end{tabular}} & \multicolumn{1}{c}{\begin{tabular}[c]{@{}c@{}}\textbf{Num. in} \\ \textbf{Chart} \end{tabular}} & \multicolumn{1}{c}{\begin{tabular}[c]{@{}c@{}}\textbf{Num. in} \\ \textbf{General}\end{tabular}} & \multicolumn{1}{c}{\cellcolor[HTML]{EFEFEF}\textbf{All}} \\ 
\midrule
LLaVA-1.6-7B          & 35.11          & 26.96          & 48.69          & 36.64          & 8.04          & \cellcolor[HTML]{F2F2F2}32.77                & 13.18                & 32.32                & 17.67                & 10.48                & \cellcolor[HTML]{F2F2F2}15.50                \\
+Majority Voting & 40.09          & 42.58          & 47.16          & 45.80          & 7.59          & \cellcolor[HTML]{F2F2F2}40.41                & 16.59                & 35.35                & 20.26                & 14.85                & \cellcolor[HTML]{F2F2F2}18.90                \\
+Chart-CoCa             & 44.78          & 56.90          & 50.66          & 49.11          & 12.50         & \cellcolor[HTML]{F2F2F2}47.83                & 20.68                & 40.40                & 24.14                & 19.21                & \cellcolor[HTML]{F2F2F2}23.10                \\
\midrule
Qwen2VL-7B            & 70.05          & 55.69          & 51.09          & 62.09          & 18.75         & \cellcolor[HTML]{F2F2F2}59.90                & 29.09                & 46.46                & 33.19                & 24.89                & \cellcolor[HTML]{F2F2F2}30.80                \\
+Majority Voting& 73.25          & 59.08          & 60.26          & 70.74          & 29.02         & \cellcolor[HTML]{F2F2F2}64.65                & 31.36                & 48.48                & 38.79                & 27.07                & \cellcolor[HTML]{F2F2F2}33.80                \\
+Chart-CoCa             & 72.66          & 78.13          & 64.41          & 76.08          & 37.05         & \cellcolor[HTML]{F2F2F2}71.75                & 34.55                & 51.52                & 41.81                & 30.13                & \cellcolor[HTML]{F2F2F2}36.90                \\
\midrule
InternVL2-8B          & 69.40          & 40.52          & 44.76          & 61.83          & 19.64         & \cellcolor[HTML]{F2F2F2}54.10                & 21.82                & 40.40                & 28.45                & 14.85                & \cellcolor[HTML]{F2F2F2}23.60                \\
+Majority Voting & 70.23          & 57.30          & 55.02          & 65.14          & 23.66         & \cellcolor[HTML]{F2F2F2}61.38                & 26.36                & 41.41                & 28.02                & 19.65                & \cellcolor[HTML]{F2F2F2}26.70                \\
+Chart-CoCa             & 74.20          & 75.46          & 58.95          & 68.19          & 26.79         & \cellcolor[HTML]{F2F2F2}69.60                & 31.59                & 44.44                & 32.33                & 25.33                & \cellcolor[HTML]{F2F2F2}31.60                \\
\midrule
InternVL2-26B         & 70.94          & 60.69          & 53.06          & 62.34          & 17.86         & \cellcolor[HTML]{F2F2F2}61.90                & 33.41                & 50.51                & 43.53                & 20.96                & \cellcolor[HTML]{F2F2F2}34.60                \\
+Majority Voting & 71.77          & 68.68          & 58.95          & 69.97          & 21.43         & \cellcolor[HTML]{F2F2F2}65.95                & 35.91                & 52.53                & 46.12                & 23.58                & \cellcolor[HTML]{F2F2F2}37.10                \\
+Chart-CoCa             & 75.09          & 79.10          & 65.07          & 74.30          & 30.80         & \cellcolor[HTML]{F2F2F2}72.63                & 37.27                & 54.55                & 48.28                & 28.38                & \cellcolor[HTML]{F2F2F2}39.50                \\
\bottomrule
\end{tabular}
\caption{The performance of our method and majority voting on different models, where the values in parentheses are the comparison with the initial model. All improvements are calculated relative to the initial model without any self-improvement strategies.}
\label{tab:diff_model}
\end{table*}

\subsection{Experiments on Different VLMs}
We evaluate our method on diverse open-source VLMs: LLaVA-1.6-Mistral-7B \citep{liu2024improved}, Qwen2VL-7B \citep{Qwen2VL}, and InternVL2 variants (8B/26B) \citep{chen2024far}. This selection covers models from both academic and industrial communities, spanning parameter scales from 7B to 26B to ensure comprehensive evaluation.

Following the standard protocol, we generate five candidate responses per query and compare majority voting with Chart-CoCa. 
As shown in Table \ref{tab:diff_model}, all models exhibit consistent improvements with majority voting, where scaling inference computation through candidate sampling boosts descriptive and reasoning performance by average 5.93 and 3.00, respectively. 
Chart-CoCa further achieves significant gains of 7.36 on descriptive tasks and 3.65 on reasoning tasks over majority voting.
Notably, the relative improvement diminishes as model size increases, with InternVL2-8B achieving 15.50/54.10=28.65\% versus 10.73/61.90=17.33\% for InternVL2-26B on descriptive tasks, suggesting our framework particularly benefits smaller models. 

\subsection{Ablation Study}
We conducted an ablation experiment to evaluate the effectiveness of each component by removing components. 
Table \ref{tab:abl} presents the overall accuracy on descriptive and reasoning tasks. 
In the $-\textit{Desc}$ setting, we eliminate the description generation module and directly generate code from the given chart.
In $-\textit{Code}$, we generate chart information from the generated chart description without code as the intermediate.
In $-\textit{Both}$, the chart information is directly generated from the given chart without both chart description and code generation.

Removing description generation ($-\textit{Desc}$) reduces performance by 4.53 and 3.60 points on descriptive and reasoning tasks, respectively. 
Eliminating code mediation ($-\textit{Code}$) causes steeper declines of 9.20 and 5.50, while the combined removal ($-\textit{Both}$) leads to maximum degradation of 13.68 and 7.10. 
This demonstrates the cumulative impact of each module, with code mediation playing a particularly critical role in preventing error accumulation during chart-to-data conversion. 
By using code as an intermediate representation, our framework ensures precise alignment between synthetic charts and their corresponding pairs of question and answer, which is essential for model self-improvement.
\begin{table*}[t]
\centering
\small
\setlength{\tabcolsep}{8pt}
\begin{tabular}{lrrrrrrrrrrr}
\toprule
\multicolumn{1}{c}{}                        & \multicolumn{6}{c}{\textbf{Descriptive Questions}}                                                                                                                                                                                                                                                     & \multicolumn{5}{c}{\textbf{Reasoning Questions}}                                                                                                                                                                                                                                                                                                                                         \\ \cmidrule(lr){2-7} \cmidrule(lr){8-12} 
\multicolumn{1}{c}{\multirow{-2}{*}{\textbf{Method}}} & \multicolumn{1}{c}{\begin{tabular}[c]{@{}c@{}}\textbf{Info.} \\ \textbf{ Extr.} \end{tabular}} & \multicolumn{1}{c}{\textbf{Enum.}} & \multicolumn{1}{c}{\begin{tabular}[c]{@{}c@{}}\textbf{Patt.}  \\ \textbf{Rec.}\end{tabular}} & \multicolumn{1}{c}{\textbf{Cntg.}} & \multicolumn{1}{c}{\textbf{Comp.}} & \multicolumn{1}{c}{\cellcolor[HTML]{EFEFEF}\textbf{All}} & \multicolumn{1}{c}{\begin{tabular}[c]{@{}c@{}}\textbf{Text in} \\ \textbf{Chart}\end{tabular}} & \multicolumn{1}{c}{\begin{tabular}[c]{@{}c@{}}\textbf{Text in} \\ \textbf{General}\end{tabular}} & \multicolumn{1}{c}{\begin{tabular}[c]{@{}c@{}}\textbf{Num. in} \\ \textbf{Chart} \end{tabular}} & \multicolumn{1}{c}{\begin{tabular}[c]{@{}c@{}}\textbf{Num. in} \\ \textbf{General}\end{tabular}} & \multicolumn{1}{c}{\cellcolor[HTML]{EFEFEF}\textbf{All}} \\ 
\midrule
Chart-CoCa             & 74.20          & 75.46          & 58.95          & 68.19          & 26.79         & \cellcolor[HTML]{F2F2F2}69.60                & 31.59                & 44.44                & 32.33                & 25.33                & \cellcolor[HTML]{F2F2F2}31.60                \\
- Desc                & 68.27          & 67.39          & 63.97          & 64.38          & 25.45         & \cellcolor[HTML]{F2F2F2}65.07                & 28.86                & 34.34                & 30.17                & 21.40                & \cellcolor[HTML]{F2F2F2}28.00                \\
- Code                & 66.25          & 60.86          & 55.46          & 60.31          & 24.11         & \cellcolor[HTML]{F2F2F2}60.40                & 26.59                & 35.35                & 27.59                & 19.65                & \cellcolor[HTML]{F2F2F2}26.10                \\
- Both                & 63.46          & 53.75          & 50.66          & 57.51          & 19.20         & \cellcolor[HTML]{F2F2F2}55.92                & 25.68                & 31.31                & 25.86                & 17.90                & \cellcolor[HTML]{F2F2F2}24.50                \\
\bottomrule
\end{tabular}
\caption{Ablation results, where the values within the parentheses represents the performance loss compared to our complete framework.}
\label{tab:abl}
\end{table*}

\begin{figure}[t]
    \centering
    \includegraphics[width=1\linewidth]{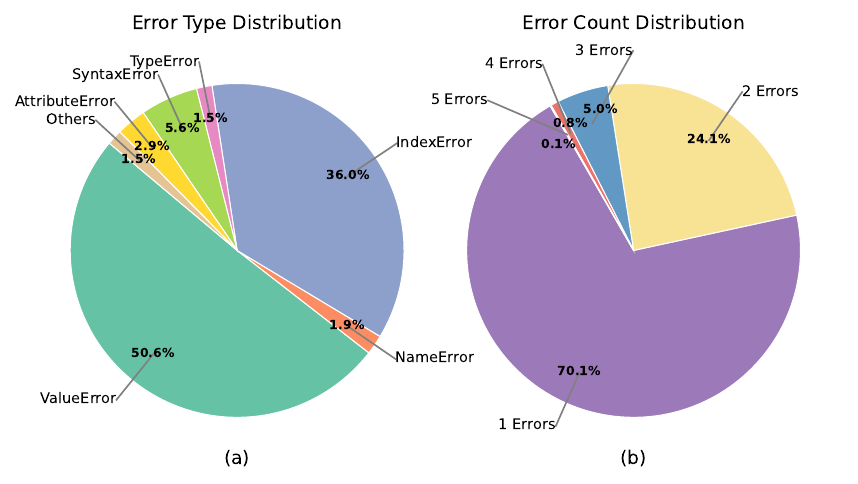}
    \caption{The left pie chart (a) illustrates the portions of different errors encountered during the code execution. The right one (b) presents the portions of single errors and multiple errors for each instance.}
    \label{fig:error-pie}
\end{figure}

\begin{figure}[t]
    \centering
    \includegraphics[width=0.98\linewidth]{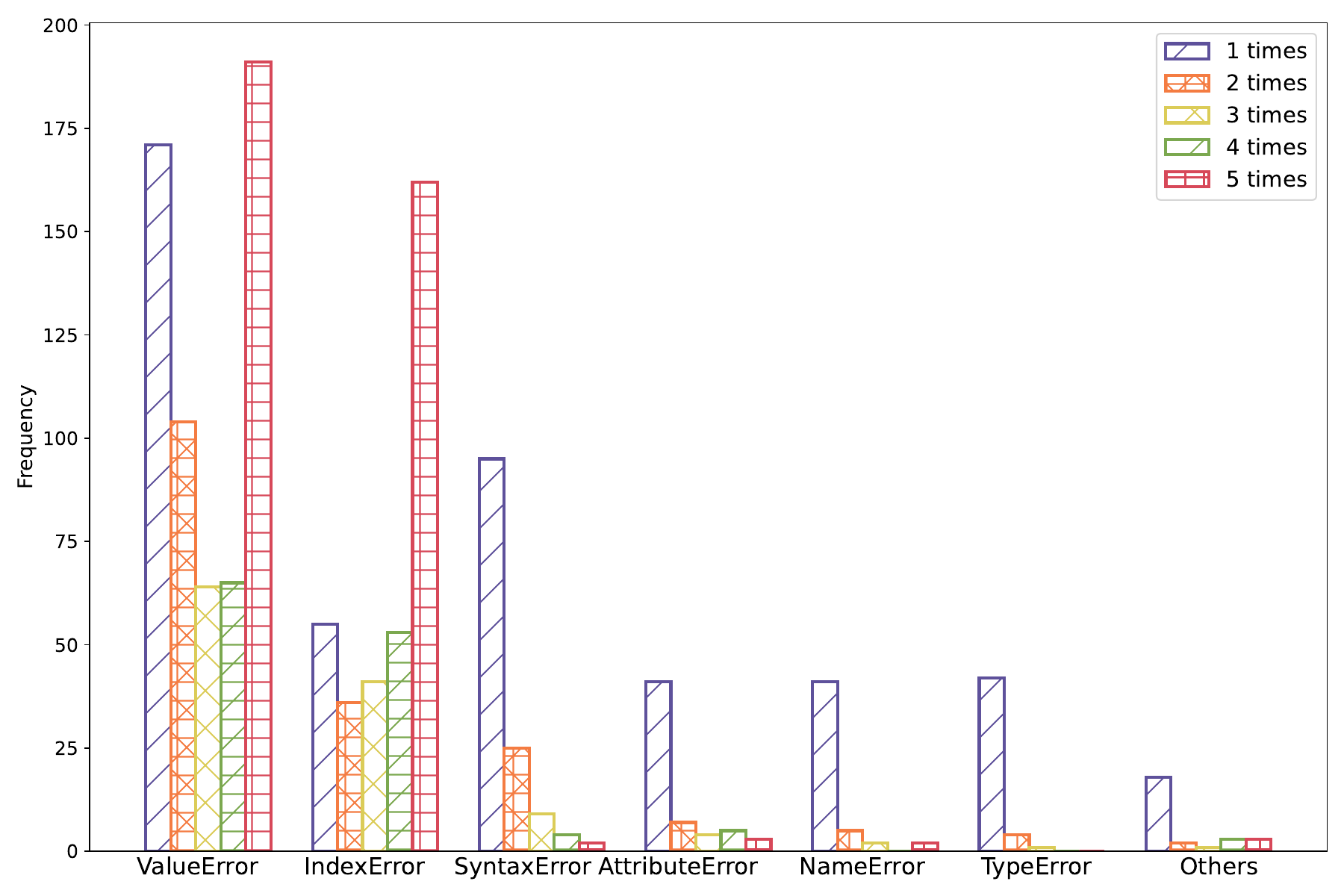}
    \caption{The frequency distribution of various error types across instances during the retries of code generation and execution.}
    \label{fig:error-bar}  
\end{figure}

\subsection{Case Study}
In addition to quantitative experiments, we also provide a case study to demonstrate how our Chart-CoCa enhances reasoning.
As observed in Table \ref{tab:case}, given a chart and a reasoning question, the five generated candidate answers include the correct answer ``Kansa City''.
However, the majority answer (``Boston'') among the five is not the correct answer.
This case shows that majority voting on the results of repeated sampling by VLMs is not always reliable.
In contrast, our method generates the final answer by taking these responses as conditions rather than directly taking one of these answers. Furthermore, the generated final answer text shows a reasoning path from an analysis of the subplot to the final answer. This generated final answer is not obtained from majority voting, but is discovered in the minority of the candidate answers through reasoning.

\begin{table}[t]
    \centering
    \begin{tabular}{p{8cm}}
    \toprule
    \includegraphics[width=0.45\textwidth]{fig/case.jpg} \\
    \midrule
    \textbf{Question.} Which city experiences the most "zig-zagging" in stay at home rates with respect to the number of daily new confirmed Covid-19 cases?  \\
    \textbf{Ground Truth.} \textbf{Kansas City}          \\
    \midrule
    \textbf{Candidate 1.} \textbf{Houston} experiences the most "zig-zagging" in stay at home rates with respect to the number of daily new confirmed Covid-19 cases. \\
    \textbf{Candidate 2.}  \textbf{Boston}. \\
    \textbf{Candidate 3.} The city that experiences the most "zig-zagging" in stay at home rates with respect to the number of daily new confirmed Covid-19 cases is \textbf{Boston}. \\
    \textbf{Candidate 4.} The city that experiences the most "zig-zagging" in stay at home rates with respect to the number of daily new confirmed Covid-19 cases is \textbf{Kansas City}. \\
    \textbf{Candidate 5.} \textbf{Boston} experiences the most "zig-zagging" in stay at home rates with respect to the number of daily new confirmed Covid-19 cases.  \\
    \midrule
    \textbf{Final Answer (Ours).} Evident from the purple line in subplot e, which shows significant fluctuations in stay-at-home rates over the course of the data presented, \textbf{Kansas City} experiences the most "zig-zagging" in stay-at-home rates with respect to the number of daily new confirmed Covid-19 cases. \\
    \bottomrule
    \end{tabular}
    \caption{A case study, where both the chart and reasoning question are derived from real data in the CharXiv dataset, and the generated answers are based on InternVL2-7B.}
    \label{tab:case}
\end{table}

\subsection{Error Analysis}
A core step in our framework is the code generation and execution. 
However, the generated code may encounter errors during execution. 
We regenerate and rerun the code with execution errors and set a maximum number of attempts (5 times of generation and execution in total). 
In this section, we conduct statistics and analysis on the execution errors throughout the process.

We calculate the proportion of each type of error, which is presented in the inner pie chart of Figure \ref{fig:error-pie}. ValueError accounts for half of the errors, followed closely by IndexError, which accounts for more than one-third. The proportions of other errors are significantly lower. SyntaxError accounts for 5.6\%, while AttributeError, TypeError, NameError and Others each account for only 1-2\%. 

The code generation and execution will be attempted up to 5 times, which may involve multiple occurrences of the same type of error or a variety of error types. 
To explore whether the model tends to repeatedly fall into a single type of error or encounter different errors, we collect the number of different error types that may occur during the multiple synthesis processes of each chart. 
We calculate the proportion of the number of error types. 
The outer ring of Figure \ref{fig:error-pie} shows the results. 
Single-type errors are the most common, accounting for 70\%. Errors of two different types occurring during the generation of the same chart account for 24\%, while the proportion of cases where the number of different error types is greater than three is less than 6\%. This indicates that despite retries, the model is prone to getting stuck in the same error.

We analyze the frequency distribution of error types. 
For instance, if a ValueError occurs twice during the generation of Chart 1, we increase the frequency count for ``2 occurrences'' by one. 
Figure \ref{fig:error-bar} presents these statistics. 
ValueError frequently occurs both once and five times, indicating that a single retry often resolves it, yet it sometimes persists even after five attempts. 
In contrast, IndexError is more likely to recur without resolution, even after multiple retries, compared to successful resolutions within four attempts. 
Targeted solutions for errors in chart code generation could enhance code execution success rates.

\section{Conclusion and Future Work}
In this study, we tackled the chart understanding challenge in VLMs. 
We proposed a method that allows VLMs to improve their chart understanding by themselves without either human-labeled data or teacher models.
Our data synthesis pipeline uses code as the intermediary to precisely connect the generated chart and the pairs of question and answer, thus obtaining training data with high correctness. 
We also explored a strategy for test-time scaling. We first trained the VLM on the synthetic data along with generated candidates for each query, to make it learn to generate an answer conditioned by candidates. 
In inference, the initial VLM first generates candidates. 
Subsequently, considering these candidates, the specifically trained VLM consolidates the final answer. 
Experiments showed that, under similar computing resources, our method has achieved the highest accuracy. 
Meanwhile, our method brought significant improvements to different VLMs, showing its adaptability.
We discuss the limitations and potential improvements, which encompasses four aspects.

\noindent \textbf{Chart-to-code Models.}
Limited by the ability of VLMs, the process of chart-to-code doesn't always meet expectations.
Recently, some chart-to-code models \citep{zhao2025chartcoder} exhibit good performance in chart-to-code tasks with a small size.
While using these external models goes against the self-improving setup, it's still exciting to employ small models to get a boost.

\noindent \textbf{Data Diversity.}
We currently synthesize charts by generating Python Matplotlib code. 
As a flexible framework, our method can be extended to other plotting libraries in the future to enhance the diversity and authenticity of charts. 
Additionally, the question-answer pairs we synthesize mainly focus on the basic elements in charts, with a relatively fixed form. 
In the future, we can expand to more flexible and diverse question-answer data.

\noindent \textbf{Reduction in Complexity.}
Our method consists of three components: data synthesis, model training, and inference.
Limited by the capabilities of current VLMs, our approach requires multiple modules and steps. 
In the future, with the support of more powerful VLMs, we may need fewer modules to achieve better results.

\noindent \textbf{Iterative Improvement.}  
Our proposed method integrates both data synthesis and model training, facilitating the exploration of iterative optimization \citep{yuan2024selfrewarding}. 
Specifically, we can: 1) employ the final-predicted data for further training; 2) utilize the well-trained model to generate data for model training in the next iteration.

\noindent \textbf{Reinforcement Learning and Process Rewarding.}  
Our methods primarily rely on supervised fine-tuning and inference scaling. Recently, reinforcement learning has shown promise in enhancing VLMs \citep{zhaifine2024}. We plan to explore this paradigm in our future endeavors.
Moreover, we propose a two-stage reasoning process. 
Some studies have used a process reward model to guide each step in the reasoning process \citep{lightman2024lets, snell2024scaling}. 
We will explore this paradigm in our future research.


\section*{Acknowledgment}
We sincerely appreciate the comments of reviewers and the area chair, which help improve this work.
This work was supported by a start-up grant at the Hong Kong University of Science and Technology (Guangzhou).

\appendix

\section{Dataset Consideration}
\label{sec:data}
We choose CharXiv \citep{wang-chartxiv-2024} as the benchmark for evaluation.
Other datasets like FigureQA \citep{kahou2017figureqa}, DVQA \citep{kafle2018dvqa}, and PlotQA \citep{methani2020plotqa}, due to their synthetic nature, cannot fully capture the complexity and diversity of real-world charts. 
ChartQA \citep{masry2022chartqa} lacks visual diversity. 
While MMC \citep{liu2024mmc}, ChartBench \citep{xu2023chartbench}, and ChartX \citep{xia2024chartx} have more real and diverse data, they have deficiencies in question types and other aspects. For example, FigureQA and others use fixed templates to pose question-answer pairs, and ChartBench generates questions according to predefined tasks.

CharXiv, on the other hand, consists of 2,323 real-world charts carefully selected from scientific papers covering 8 major disciplines on arXiv. 
By designing descriptive questions (requiring an understanding of basic chart information such as titles, labels, and ticks) and reasoning questions (needing comparisons, approximations, and fine-grained analysis), it clearly differentiates between visual understanding and reasoning. 
All questions are carefully curated by human experts, and all ground-truth answers are manually verified, making it a high-quality dataset. 
It also adopts a short-answer format that is suitable for automatic scoring based on LLMs.

\section{GenAI Usage Disclosure}
We use GPT-4o and DeepSeek for the purpose of correcting grammar and enhancing expressions.
In our research, we employed generative models for code generation and data synthesis, which constitute one of our central focuses aiming at enhancing model capabilities through synthetic data.
However, it is crucial to clarify that: (1) The experimental code implementations were developed independently, (2) All experimental datasets were derived from empirical results.

\bibliographystyle{ACM-Reference-Format}
\bibliography{sample-base}


\begin{thebibliography}{47}


\ifx \showCODEN    \undefined \def \showCODEN     #1{\unskip}     \fi
\ifx \showISBNx    \undefined \def \showISBNx     #1{\unskip}     \fi
\ifx \showISBNxiii \undefined \def \showISBNxiii  #1{\unskip}     \fi
\ifx \showISSN     \undefined \def \showISSN      #1{\unskip}     \fi
\ifx \showLCCN     \undefined \def \showLCCN      #1{\unskip}     \fi
\ifx \shownote     \undefined \def \shownote      #1{#1}          \fi
\ifx \showarticletitle \undefined \def \showarticletitle #1{#1}   \fi
\ifx \showURL      \undefined \def \showURL       {\relax}        \fi
\providecommand\bibfield[2]{#2}
\providecommand\bibinfo[2]{#2}
\providecommand\natexlab[1]{#1}
\providecommand\showeprint[2][]{arXiv:#2}

\bibitem[Achiam et~al\mbox{.}(2023)]%
        {achiam2023gpt}
\bibfield{author}{\bibinfo{person}{Josh Achiam}, \bibinfo{person}{Steven Adler}, \bibinfo{person}{Sandhini Agarwal}, \bibinfo{person}{Lama Ahmad}, \bibinfo{person}{Ilge Akkaya}, \bibinfo{person}{Florencia~Leoni Aleman}, \bibinfo{person}{Diogo Almeida}, \bibinfo{person}{Janko Altenschmidt}, \bibinfo{person}{Sam Altman}, \bibinfo{person}{Shyamal Anadkat}, {et~al\mbox{.}}} \bibinfo{year}{2023}\natexlab{}.
\newblock \showarticletitle{Gpt-4 technical report}.
\newblock \bibinfo{journal}{\emph{arXiv preprint arXiv:2303.08774}} (\bibinfo{year}{2023}).
\newblock


\bibitem[Amini et~al\mbox{.}(2025)]%
        {amini2025self}
\bibfield{author}{\bibinfo{person}{Massih-Reza Amini}, \bibinfo{person}{Vasilii Feofanov}, \bibinfo{person}{Loic Pauletto}, \bibinfo{person}{Lies Hadjadj}, \bibinfo{person}{Emilie Devijver}, {and} \bibinfo{person}{Yury Maximov}.} \bibinfo{year}{2025}\natexlab{}.
\newblock \showarticletitle{Self-training: A survey}.
\newblock \bibinfo{journal}{\emph{Neurocomputing}}  \bibinfo{volume}{616} (\bibinfo{year}{2025}), \bibinfo{pages}{128904}.
\newblock


\bibitem[Bai et~al\mbox{.}(2023)]%
        {bai2023qwen}
\bibfield{author}{\bibinfo{person}{Jinze Bai}, \bibinfo{person}{Shuai Bai}, \bibinfo{person}{Shusheng Yang}, \bibinfo{person}{Shijie Wang}, \bibinfo{person}{Sinan Tan}, \bibinfo{person}{Peng Wang}, \bibinfo{person}{Junyang Lin}, \bibinfo{person}{Chang Zhou}, {and} \bibinfo{person}{Jingren Zhou}.} \bibinfo{year}{2023}\natexlab{}.
\newblock \showarticletitle{Qwen-vl: A frontier large vision-language model with versatile abilities}.
\newblock \bibinfo{journal}{\emph{arXiv preprint arXiv:2308.12966}} (\bibinfo{year}{2023}).
\newblock


\bibitem[Bansal et~al\mbox{.}(2024)]%
        {bansal2024smaller}
\bibfield{author}{\bibinfo{person}{Hritik Bansal}, \bibinfo{person}{Arian Hosseini}, \bibinfo{person}{Rishabh Agarwal}, \bibinfo{person}{Vinh~Q. Tran}, {and} \bibinfo{person}{Mehran Kazemi}.} \bibinfo{year}{2024}\natexlab{}.
\newblock \showarticletitle{Smaller, Weaker, Yet Better: Training {LLM} Reasoners via Compute-Optimal Sampling}. In \bibinfo{booktitle}{\emph{The 4th Workshop on Mathematical Reasoning and AI at NeurIPS'24}}.
\newblock
\urldef\tempurl%
\url{https://openreview.net/forum?id=HuYSURUxs2}
\showURL{%
\tempurl}


\bibitem[Chen et~al\mbox{.}(2024a)]%
        {chen2024selfplay}
\bibfield{author}{\bibinfo{person}{Zixiang Chen}, \bibinfo{person}{Yihe Deng}, \bibinfo{person}{Huizhuo Yuan}, \bibinfo{person}{Kaixuan Ji}, {and} \bibinfo{person}{Quanquan Gu}.} \bibinfo{year}{2024}\natexlab{a}.
\newblock \showarticletitle{Self-Play Fine-Tuning Converts Weak Language Models to Strong Language Models}. In \bibinfo{booktitle}{\emph{ICML 2024}}.
\newblock


\bibitem[Chen et~al\mbox{.}(2024b)]%
        {chen2024self}
\bibfield{author}{\bibinfo{person}{Zixiang Chen}, \bibinfo{person}{Yihe Deng}, \bibinfo{person}{Huizhuo Yuan}, \bibinfo{person}{Kaixuan Ji}, {and} \bibinfo{person}{Quanquan Gu}.} \bibinfo{year}{2024}\natexlab{b}.
\newblock \showarticletitle{Self-Play Fine-Tuning Converts Weak Language Models to Strong Language Models}. In \bibinfo{booktitle}{\emph{Forty-first International Conference on Machine Learning, {ICML} 2024, Vienna, Austria, July 21-27, 2024}}. \bibinfo{publisher}{OpenReview.net}.
\newblock
\urldef\tempurl%
\url{https://openreview.net/forum?id=O4cHTxW9BS}
\showURL{%
\tempurl}


\bibitem[Chen et~al\mbox{.}(2024c)]%
        {chen2024far}
\bibfield{author}{\bibinfo{person}{Zhe Chen}, \bibinfo{person}{Weiyun Wang}, \bibinfo{person}{Hao Tian}, \bibinfo{person}{Shenglong Ye}, \bibinfo{person}{Zhangwei Gao}, \bibinfo{person}{Erfei Cui}, \bibinfo{person}{Wenwen Tong}, \bibinfo{person}{Kongzhi Hu}, \bibinfo{person}{Jiapeng Luo}, \bibinfo{person}{Zheng Ma}, {et~al\mbox{.}}} \bibinfo{year}{2024}\natexlab{c}.
\newblock \showarticletitle{How far are we to gpt-4v? closing the gap to commercial multimodal models with open-source suites}.
\newblock \bibinfo{journal}{\emph{arXiv preprint arXiv:2404.16821}} (\bibinfo{year}{2024}).
\newblock


\bibitem[Cobbe et~al\mbox{.}(2021)]%
        {cobbe2021training}
\bibfield{author}{\bibinfo{person}{Karl Cobbe}, \bibinfo{person}{Vineet Kosaraju}, \bibinfo{person}{Mohammad Bavarian}, \bibinfo{person}{Mark Chen}, \bibinfo{person}{Heewoo Jun}, \bibinfo{person}{Lukasz Kaiser}, \bibinfo{person}{Matthias Plappert}, \bibinfo{person}{Jerry Tworek}, \bibinfo{person}{Jacob Hilton}, \bibinfo{person}{Reiichiro Nakano}, {et~al\mbox{.}}} \bibinfo{year}{2021}\natexlab{}.
\newblock \showarticletitle{Training verifiers to solve math word problems}.
\newblock \bibinfo{journal}{\emph{arXiv preprint arXiv:2110.14168}} (\bibinfo{year}{2021}).
\newblock


\bibitem[Dai et~al\mbox{.}(2023)]%
        {dai2023instructblip}
\bibfield{author}{\bibinfo{person}{Wenliang Dai}, \bibinfo{person}{Junnan Li}, \bibinfo{person}{Dongxu Li}, \bibinfo{person}{Anthony Tiong}, \bibinfo{person}{Junqi Zhao}, \bibinfo{person}{Weisheng Wang}, \bibinfo{person}{Boyang Li}, \bibinfo{person}{Pascale Fung}, {and} \bibinfo{person}{Steven Hoi}.} \bibinfo{year}{2023}\natexlab{}.
\newblock \showarticletitle{Instruct{BLIP}: Towards General-purpose Vision-Language Models with Instruction Tuning}. In \bibinfo{booktitle}{\emph{Thirty-seventh Conference on Neural Information Processing Systems}}.
\newblock
\urldef\tempurl%
\url{https://openreview.net/forum?id=vvoWPYqZJA}
\showURL{%
\tempurl}


\bibitem[Deng et~al\mbox{.}(2024)]%
        {deng2024rephraserespondletlarge}
\bibfield{author}{\bibinfo{person}{Yihe Deng}, \bibinfo{person}{Weitong Zhang}, \bibinfo{person}{Zixiang Chen}, {and} \bibinfo{person}{Quanquan Gu}.} \bibinfo{year}{2024}\natexlab{}.
\newblock \bibinfo{title}{Rephrase and Respond: Let Large Language Models Ask Better Questions for Themselves}.
\newblock
\showeprint[arxiv]{2311.04205}~[cs.CL]
\urldef\tempurl%
\url{https://arxiv.org/abs/2311.04205}
\showURL{%
\tempurl}


\bibitem[Dong et~al\mbox{.}(2024)]%
        {dong2024internlm}
\bibfield{author}{\bibinfo{person}{Xiaoyi Dong}, \bibinfo{person}{Pan Zhang}, \bibinfo{person}{Yuhang Zang}, \bibinfo{person}{Yuhang Cao}, \bibinfo{person}{Bin Wang}, \bibinfo{person}{Linke Ouyang}, \bibinfo{person}{Xilin Wei}, \bibinfo{person}{Songyang Zhang}, \bibinfo{person}{Haodong Duan}, \bibinfo{person}{Maosong Cao}, {et~al\mbox{.}}} \bibinfo{year}{2024}\natexlab{}.
\newblock \showarticletitle{Internlm-xcomposer2: Mastering free-form text-image composition and comprehension in vision-language large model}.
\newblock \bibinfo{journal}{\emph{arXiv preprint arXiv:2401.16420}} (\bibinfo{year}{2024}).
\newblock


\bibitem[Gudivada et~al\mbox{.}(2017)]%
        {gudivada2017data}
\bibfield{author}{\bibinfo{person}{Venkat Gudivada}, \bibinfo{person}{Amy Apon}, {and} \bibinfo{person}{Junhua Ding}.} \bibinfo{year}{2017}\natexlab{}.
\newblock \showarticletitle{Data quality considerations for big data and machine learning: Going beyond data cleaning and transformations}.
\newblock \bibinfo{journal}{\emph{International Journal on Advances in Software}} \bibinfo{volume}{10}, \bibinfo{number}{1} (\bibinfo{year}{2017}), \bibinfo{pages}{1--20}.
\newblock


\bibitem[Guo et~al\mbox{.}(2025)]%
        {guo2025deepseek}
\bibfield{author}{\bibinfo{person}{Daya Guo}, \bibinfo{person}{Dejian Yang}, \bibinfo{person}{Haowei Zhang}, \bibinfo{person}{Junxiao Song}, \bibinfo{person}{Ruoyu Zhang}, \bibinfo{person}{Runxin Xu}, \bibinfo{person}{Qihao Zhu}, \bibinfo{person}{Shirong Ma}, \bibinfo{person}{Peiyi Wang}, \bibinfo{person}{Xiao Bi}, {et~al\mbox{.}}} \bibinfo{year}{2025}\natexlab{}.
\newblock \showarticletitle{Deepseek-r1: Incentivizing reasoning capability in llms via reinforcement learning}.
\newblock \bibinfo{journal}{\emph{arXiv preprint arXiv:2501.12948}} (\bibinfo{year}{2025}).
\newblock


\bibitem[He et~al\mbox{.}(2024a)]%
        {he2024self}
\bibfield{author}{\bibinfo{person}{Jiayi He}, \bibinfo{person}{Hehai Lin}, \bibinfo{person}{Qingyun Wang}, \bibinfo{person}{Yi Fung}, {and} \bibinfo{person}{Heng Ji}.} \bibinfo{year}{2024}\natexlab{a}.
\newblock \showarticletitle{Self-Correction is More than Refinement: {A} Learning Framework for Visual and Language Reasoning Tasks}.
\newblock \bibinfo{journal}{\emph{CoRR}}  \bibinfo{volume}{abs/2410.04055} (\bibinfo{year}{2024}).
\newblock
\href{https://doi.org/10.48550/ARXIV.2410.04055}{doi:\nolinkurl{10.48550/ARXIV.2410.04055}}
\showeprint[arXiv]{2410.04055}


\bibitem[He et~al\mbox{.}(2024b)]%
        {he2024distill}
\bibfield{author}{\bibinfo{person}{Wei He}, \bibinfo{person}{Zhiheng Xi}, \bibinfo{person}{Wanxu Zhao}, \bibinfo{person}{Xiaoran Fan}, \bibinfo{person}{Yiwen Ding}, \bibinfo{person}{Zifei Shan}, \bibinfo{person}{Tao Gui}, \bibinfo{person}{Qi Zhang}, {and} \bibinfo{person}{Xuanjing Huang}.} \bibinfo{year}{2024}\natexlab{b}.
\newblock \showarticletitle{Distill Visual Chart Reasoning Ability from LLMs to MLLMs}.
\newblock \bibinfo{journal}{\emph{arXiv preprint arXiv:2410.18798}} (\bibinfo{year}{2024}).
\newblock


\bibitem[Kafle et~al\mbox{.}(2018)]%
        {kafle2018dvqa}
\bibfield{author}{\bibinfo{person}{Kushal Kafle}, \bibinfo{person}{Brian Price}, \bibinfo{person}{Scott Cohen}, {and} \bibinfo{person}{Christopher Kanan}.} \bibinfo{year}{2018}\natexlab{}.
\newblock \showarticletitle{Dvqa: Understanding data visualizations via question answering}. In \bibinfo{booktitle}{\emph{Proceedings of the IEEE conference on computer vision and pattern recognition}}. \bibinfo{pages}{5648--5656}.
\newblock


\bibitem[Kahou et~al\mbox{.}(2017)]%
        {kahou2017figureqa}
\bibfield{author}{\bibinfo{person}{Samira~Ebrahimi Kahou}, \bibinfo{person}{Vincent Michalski}, \bibinfo{person}{Adam Atkinson}, \bibinfo{person}{{\'A}kos K{\'a}d{\'a}r}, \bibinfo{person}{Adam Trischler}, {and} \bibinfo{person}{Yoshua Bengio}.} \bibinfo{year}{2017}\natexlab{}.
\newblock \showarticletitle{Figureqa: An annotated figure dataset for visual reasoning}.
\newblock \bibinfo{journal}{\emph{arXiv preprint arXiv:1710.07300}} (\bibinfo{year}{2017}).
\newblock


\bibitem[Lewkowycz et~al\mbox{.}(2022)]%
        {lewkowycz2022solving}
\bibfield{author}{\bibinfo{person}{Aitor Lewkowycz}, \bibinfo{person}{Anders Andreassen}, \bibinfo{person}{David Dohan}, \bibinfo{person}{Ethan Dyer}, \bibinfo{person}{Henryk Michalewski}, \bibinfo{person}{Vinay Ramasesh}, \bibinfo{person}{Ambrose Slone}, \bibinfo{person}{Cem Anil}, \bibinfo{person}{Imanol Schlag}, \bibinfo{person}{Theo Gutman-Solo}, {et~al\mbox{.}}} \bibinfo{year}{2022}\natexlab{}.
\newblock \showarticletitle{Solving quantitative reasoning problems with language models}.
\newblock \bibinfo{journal}{\emph{Advances in Neural Information Processing Systems}}  \bibinfo{volume}{35} (\bibinfo{year}{2022}), \bibinfo{pages}{3843--3857}.
\newblock


\bibitem[Lightman et~al\mbox{.}(2024)]%
        {lightman2024lets}
\bibfield{author}{\bibinfo{person}{Hunter Lightman}, \bibinfo{person}{Vineet Kosaraju}, \bibinfo{person}{Yuri Burda}, \bibinfo{person}{Harrison Edwards}, \bibinfo{person}{Bowen Baker}, \bibinfo{person}{Teddy Lee}, \bibinfo{person}{Jan Leike}, \bibinfo{person}{John Schulman}, \bibinfo{person}{Ilya Sutskever}, {and} \bibinfo{person}{Karl Cobbe}.} \bibinfo{year}{2024}\natexlab{}.
\newblock \showarticletitle{Let's Verify Step by Step}. In \bibinfo{booktitle}{\emph{The Twelfth International Conference on Learning Representations}}. \bibinfo{publisher}{ICLR}, \bibinfo{address}{Vienna, Austria}.
\newblock
\urldef\tempurl%
\url{https://openreview.net/forum?id=v8L0pN6EOi}
\showURL{%
\tempurl}


\bibitem[Liu et~al\mbox{.}(2024b)]%
        {liu2024mmc}
\bibfield{author}{\bibinfo{person}{Fuxiao Liu}, \bibinfo{person}{Xiaoyang Wang}, \bibinfo{person}{Wenlin Yao}, \bibinfo{person}{Jianshu Chen}, \bibinfo{person}{Kaiqiang Song}, \bibinfo{person}{Sangwoo Cho}, \bibinfo{person}{Yaser Yacoob}, {and} \bibinfo{person}{Dong Yu}.} \bibinfo{year}{2024}\natexlab{b}.
\newblock \showarticletitle{MMC: Advancing Multimodal Chart Understanding with Large-scale Instruction Tuning}. In \bibinfo{booktitle}{\emph{Proceedings of the 2024 Conference of the North American Chapter of the Association for Computational Linguistics: Human Language Technologies (Volume 1: Long Papers)}}. \bibinfo{publisher}{ACL}, \bibinfo{pages}{1287--1310}.
\newblock


\bibitem[Liu et~al\mbox{.}(2024a)]%
        {liu2024improved}
\bibfield{author}{\bibinfo{person}{Haotian Liu}, \bibinfo{person}{Chunyuan Li}, \bibinfo{person}{Yuheng Li}, {and} \bibinfo{person}{Yong~Jae Lee}.} \bibinfo{year}{2024}\natexlab{a}.
\newblock \showarticletitle{Improved Baselines with Visual Instruction Tuning}. In \bibinfo{booktitle}{\emph{{IEEE/CVF} Conference on Computer Vision and Pattern Recognition, {CVPR} 2024, Seattle, WA, USA, June 16-22, 2024}}. \bibinfo{publisher}{{IEEE}}, \bibinfo{pages}{26286--26296}.
\newblock
\href{https://doi.org/10.1109/CVPR52733.2024.02484}{doi:\nolinkurl{10.1109/CVPR52733.2024.02484}}


\bibitem[Long et~al\mbox{.}(2024)]%
        {long-etal-2024-llms}
\bibfield{author}{\bibinfo{person}{Lin Long}, \bibinfo{person}{Rui Wang}, \bibinfo{person}{Ruixuan Xiao}, \bibinfo{person}{Junbo Zhao}, \bibinfo{person}{Xiao Ding}, \bibinfo{person}{Gang Chen}, {and} \bibinfo{person}{Haobo Wang}.} \bibinfo{year}{2024}\natexlab{}.
\newblock \showarticletitle{On {LLM}s-Driven Synthetic Data Generation, Curation, and Evaluation: A Survey}. In \bibinfo{booktitle}{\emph{Findings of the Association for Computational Linguistics ACL 2024}}, \bibfield{editor}{\bibinfo{person}{Lun-Wei Ku}, \bibinfo{person}{Andre Martins}, {and} \bibinfo{person}{Vivek Srikumar}} (Eds.). \bibinfo{publisher}{Association for Computational Linguistics}, \bibinfo{address}{Bangkok, Thailand and virtual meeting}, \bibinfo{pages}{11065--11082}.
\newblock
\href{https://doi.org/10.18653/v1/2024.findings-acl.658}{doi:\nolinkurl{10.18653/v1/2024.findings-acl.658}}


\bibitem[Madaan et~al\mbox{.}(2023)]%
        {aman-selfrefine-2023}
\bibfield{author}{\bibinfo{person}{Aman Madaan}, \bibinfo{person}{Niket Tandon}, \bibinfo{person}{Prakhar Gupta}, \bibinfo{person}{Skyler Hallinan}, \bibinfo{person}{Luyu Gao}, \bibinfo{person}{Sarah Wiegreffe}, \bibinfo{person}{Uri Alon}, \bibinfo{person}{Nouha Dziri}, \bibinfo{person}{Shrimai Prabhumoye}, \bibinfo{person}{Yiming Yang}, \bibinfo{person}{Shashank Gupta}, \bibinfo{person}{Bodhisattwa~Prasad Majumder}, \bibinfo{person}{Katherine Hermann}, \bibinfo{person}{Sean Welleck}, \bibinfo{person}{Amir Yazdanbakhsh}, {and} \bibinfo{person}{Peter Clark}.} \bibinfo{year}{2023}\natexlab{}.
\newblock \showarticletitle{Self-Refine: Iterative Refinement with Self-Feedback}. In \bibinfo{booktitle}{\emph{Advances in Neural Information Processing Systems}}, \bibfield{editor}{\bibinfo{person}{A.~Oh}, \bibinfo{person}{T.~Naumann}, \bibinfo{person}{A.~Globerson}, \bibinfo{person}{K.~Saenko}, \bibinfo{person}{M.~Hardt}, {and} \bibinfo{person}{S.~Levine}} (Eds.), Vol.~\bibinfo{volume}{36}. \bibinfo{publisher}{Curran Associates, Inc.}, \bibinfo{pages}{46534--46594}.
\newblock
\urldef\tempurl%
\url{https://proceedings.neurips.cc/paper_files/paper/2023/file/91edff07232fb1b55a505a9e9f6c0ff3-Paper-Conference.pdf}
\showURL{%
\tempurl}


\bibitem[Masry et~al\mbox{.}(2022)]%
        {masry2022chartqa}
\bibfield{author}{\bibinfo{person}{Ahmed Masry}, \bibinfo{person}{Do~Xuan Long}, \bibinfo{person}{Jia~Qing Tan}, \bibinfo{person}{Shafiq Joty}, {and} \bibinfo{person}{Enamul Hoque}.} \bibinfo{year}{2022}\natexlab{}.
\newblock \showarticletitle{{C}hart{QA}: A Benchmark for Question Answering about Charts with Visual and Logical Reasoning}. In \bibinfo{booktitle}{\emph{Findings of the Association for Computational Linguistics: ACL 2022}}, \bibfield{editor}{\bibinfo{person}{Smaranda Muresan}, \bibinfo{person}{Preslav Nakov}, {and} \bibinfo{person}{Aline Villavicencio}} (Eds.). \bibinfo{publisher}{Association for Computational Linguistics}, \bibinfo{address}{Dublin, Ireland}, \bibinfo{pages}{2263--2279}.
\newblock
\href{https://doi.org/10.18653/v1/2022.findings-acl.177}{doi:\nolinkurl{10.18653/v1/2022.findings-acl.177}}


\bibitem[Masry et~al\mbox{.}(2024)]%
        {masry2024chartinstruct}
\bibfield{author}{\bibinfo{person}{Ahmed Masry}, \bibinfo{person}{Mehrad Shahmohammadi}, \bibinfo{person}{Md~Rizwan Parvez}, \bibinfo{person}{Enamul Hoque}, {and} \bibinfo{person}{Shafiq Joty}.} \bibinfo{year}{2024}\natexlab{}.
\newblock \showarticletitle{{C}hart{I}nstruct: Instruction Tuning for Chart Comprehension and Reasoning}. In \bibinfo{booktitle}{\emph{Findings of the Association for Computational Linguistics: ACL 2024}}, \bibfield{editor}{\bibinfo{person}{Lun-Wei Ku}, \bibinfo{person}{Andre Martins}, {and} \bibinfo{person}{Vivek Srikumar}} (Eds.). \bibinfo{publisher}{Association for Computational Linguistics}, \bibinfo{address}{Bangkok, Thailand}, \bibinfo{pages}{10387--10409}.
\newblock
\href{https://doi.org/10.18653/v1/2024.findings-acl.619}{doi:\nolinkurl{10.18653/v1/2024.findings-acl.619}}


\bibitem[Meng et~al\mbox{.}(2024)]%
        {meng2024chartassisstant}
\bibfield{author}{\bibinfo{person}{Fanqing Meng}, \bibinfo{person}{Wenqi Shao}, \bibinfo{person}{Quanfeng Lu}, \bibinfo{person}{Peng Gao}, \bibinfo{person}{Kaipeng Zhang}, \bibinfo{person}{Yu Qiao}, {and} \bibinfo{person}{Ping Luo}.} \bibinfo{year}{2024}\natexlab{}.
\newblock \showarticletitle{Chartassisstant: A universal chart multimodal language model via chart-to-table pre-training and multitask instruction tuning}.
\newblock \bibinfo{journal}{\emph{arXiv preprint arXiv:2401.02384}} (\bibinfo{year}{2024}).
\newblock


\bibitem[Methani et~al\mbox{.}(2020)]%
        {methani2020plotqa}
\bibfield{author}{\bibinfo{person}{Nitesh Methani}, \bibinfo{person}{Pritha Ganguly}, \bibinfo{person}{Mitesh~M Khapra}, {and} \bibinfo{person}{Pratyush Kumar}.} \bibinfo{year}{2020}\natexlab{}.
\newblock \showarticletitle{Plotqa: Reasoning over scientific plots}. In \bibinfo{booktitle}{\emph{Proceedings of the IEEE/CVF Winter Conference on Applications of Computer Vision}}. \bibinfo{pages}{1527--1536}.
\newblock


\bibitem[Patel et~al\mbox{.}(2024)]%
        {patel-etal-2024-datadreamer}
\bibfield{author}{\bibinfo{person}{Ajay Patel}, \bibinfo{person}{Colin Raffel}, {and} \bibinfo{person}{Chris Callison-Burch}.} \bibinfo{year}{2024}\natexlab{}.
\newblock \showarticletitle{{D}ata{D}reamer: A Tool for Synthetic Data Generation and Reproducible {LLM} Workflows}. In \bibinfo{booktitle}{\emph{Proceedings of the 62nd Annual Meeting of the Association for Computational Linguistics (Volume 1: Long Papers)}}, \bibfield{editor}{\bibinfo{person}{Lun-Wei Ku}, \bibinfo{person}{Andre Martins}, {and} \bibinfo{person}{Vivek Srikumar}} (Eds.). \bibinfo{publisher}{Association for Computational Linguistics}, \bibinfo{address}{Bangkok, Thailand}, \bibinfo{pages}{3781--3799}.
\newblock
\href{https://doi.org/10.18653/v1/2024.acl-long.208}{doi:\nolinkurl{10.18653/v1/2024.acl-long.208}}


\bibitem[Polyzotis et~al\mbox{.}(2018)]%
        {polyzotis2018data}
\bibfield{author}{\bibinfo{person}{Neoklis Polyzotis}, \bibinfo{person}{Sudip Roy}, \bibinfo{person}{Steven~Euijong Whang}, {and} \bibinfo{person}{Martin Zinkevich}.} \bibinfo{year}{2018}\natexlab{}.
\newblock \showarticletitle{Data lifecycle challenges in production machine learning: a survey}.
\newblock \bibinfo{journal}{\emph{ACM SIGMOD Record}} \bibinfo{volume}{47}, \bibinfo{number}{2} (\bibinfo{year}{2018}), \bibinfo{pages}{17--28}.
\newblock


\bibitem[Rajbhandari et~al\mbox{.}(2020)]%
        {rajbhandari2020zero}
\bibfield{author}{\bibinfo{person}{Samyam Rajbhandari}, \bibinfo{person}{Jeff Rasley}, \bibinfo{person}{Olatunji Ruwase}, {and} \bibinfo{person}{Yuxiong He}.} \bibinfo{year}{2020}\natexlab{}.
\newblock \showarticletitle{ZeRO: memory optimizations toward training trillion parameter models}. In \bibinfo{booktitle}{\emph{Proceedings of the International Conference for High Performance Computing, Networking, Storage and Analysis, {SC} 2020, Virtual Event / Atlanta, Georgia, USA, November 9-19, 2020}}, \bibfield{editor}{\bibinfo{person}{Christine Cuicchi}, \bibinfo{person}{Irene Qualters}, {and} \bibinfo{person}{William~T. Kramer}} (Eds.). \bibinfo{publisher}{{IEEE/ACM}}, \bibinfo{pages}{20}.
\newblock
\href{https://doi.org/10.1109/SC41405.2020.00024}{doi:\nolinkurl{10.1109/SC41405.2020.00024}}


\bibitem[Snell et~al\mbox{.}(2024)]%
        {snell2024scaling}
\bibfield{author}{\bibinfo{person}{Charlie Snell}, \bibinfo{person}{Jaehoon Lee}, \bibinfo{person}{Kelvin Xu}, {and} \bibinfo{person}{Aviral Kumar}.} \bibinfo{year}{2024}\natexlab{}.
\newblock \showarticletitle{Scaling llm test-time compute optimally can be more effective than scaling model parameters}.
\newblock \bibinfo{journal}{\emph{arXiv preprint arXiv:2408.03314}} (\bibinfo{year}{2024}).
\newblock


\bibitem[Taori et~al\mbox{.}(2023)]%
        {alpaca}
\bibfield{author}{\bibinfo{person}{Rohan Taori}, \bibinfo{person}{Ishaan Gulrajani}, \bibinfo{person}{Tianyi Zhang}, \bibinfo{person}{Yann Dubois}, \bibinfo{person}{Xuechen Li}, \bibinfo{person}{Carlos Guestrin}, \bibinfo{person}{Percy Liang}, {and} \bibinfo{person}{Tatsunori~B. Hashimoto}.} \bibinfo{year}{2023}\natexlab{}.
\newblock \bibinfo{title}{Stanford Alpaca: An Instruction-following LLaMA model}.
\newblock \bibinfo{howpublished}{\url{https://github.com/tatsu-lab/stanford_alpaca}}.
\newblock


\bibitem[Wang et~al\mbox{.}(2024a)]%
        {Qwen2VL}
\bibfield{author}{\bibinfo{person}{Peng Wang}, \bibinfo{person}{Shuai Bai}, \bibinfo{person}{Sinan Tan}, \bibinfo{person}{Shijie Wang}, \bibinfo{person}{Zhihao Fan}, \bibinfo{person}{Jinze Bai}, \bibinfo{person}{Keqin Chen}, \bibinfo{person}{Xuejing Liu}, \bibinfo{person}{Jialin Wang}, \bibinfo{person}{Wenbin Ge}, \bibinfo{person}{Yang Fan}, \bibinfo{person}{Kai Dang}, \bibinfo{person}{Mengfei Du}, \bibinfo{person}{Xuancheng Ren}, \bibinfo{person}{Rui Men}, \bibinfo{person}{Dayiheng Liu}, \bibinfo{person}{Chang Zhou}, \bibinfo{person}{Jingren Zhou}, {and} \bibinfo{person}{Junyang Lin}.} \bibinfo{year}{2024}\natexlab{a}.
\newblock \showarticletitle{Qwen2-VL: Enhancing Vision-Language Model's Perception of the World at Any Resolution}.
\newblock \bibinfo{journal}{\emph{arXiv preprint arXiv:2409.12191}} (\bibinfo{year}{2024}).
\newblock


\bibitem[Wang et~al\mbox{.}(2024b)]%
        {wang2024enhancing}
\bibfield{author}{\bibinfo{person}{Xiyao Wang}, \bibinfo{person}{Jiuhai Chen}, \bibinfo{person}{Zhaoyang Wang}, \bibinfo{person}{Yuhang Zhou}, \bibinfo{person}{Yiyang Zhou}, \bibinfo{person}{Huaxiu Yao}, \bibinfo{person}{Tianyi Zhou}, \bibinfo{person}{Tom Goldstein}, \bibinfo{person}{Parminder Bhatia}, \bibinfo{person}{Furong Huang}, {and} \bibinfo{person}{Cao Xiao}.} \bibinfo{year}{2024}\natexlab{b}.
\newblock \showarticletitle{Enhancing Visual-Language Modality Alignment in Large Vision Language Models via Self-Improvement}.
\newblock \bibinfo{journal}{\emph{CoRR}}  \bibinfo{volume}{abs/2405.15973} (\bibinfo{year}{2024}).
\newblock
\href{https://doi.org/10.48550/ARXIV.2405.15973}{doi:\nolinkurl{10.48550/ARXIV.2405.15973}}
\showeprint[arXiv]{2405.15973}


\bibitem[Wang et~al\mbox{.}(2023b)]%
        {wangself2023}
\bibfield{author}{\bibinfo{person}{Xuezhi Wang}, \bibinfo{person}{Jason Wei}, \bibinfo{person}{Dale Schuurmans}, \bibinfo{person}{Quoc~V Le}, \bibinfo{person}{Ed~H Chi}, \bibinfo{person}{Sharan Narang}, \bibinfo{person}{Aakanksha Chowdhery}, {and} \bibinfo{person}{Denny Zhou}.} \bibinfo{year}{2023}\natexlab{b}.
\newblock \showarticletitle{Self-Consistency Improves Chain of Thought Reasoning in Language Models}. In \bibinfo{booktitle}{\emph{The Eleventh International Conference on Learning Representations}}.
\newblock


\bibitem[Wang et~al\mbox{.}(2023a)]%
        {wang2023self}
\bibfield{author}{\bibinfo{person}{Yizhong Wang}, \bibinfo{person}{Yeganeh Kordi}, \bibinfo{person}{Swaroop Mishra}, \bibinfo{person}{Alisa Liu}, \bibinfo{person}{Noah~A Smith}, \bibinfo{person}{Daniel Khashabi}, {and} \bibinfo{person}{Hannaneh Hajishirzi}.} \bibinfo{year}{2023}\natexlab{a}.
\newblock \showarticletitle{Self-Instruct: Aligning Language Models with Self-Generated Instructions}. In \bibinfo{booktitle}{\emph{The 61st Annual Meeting Of The Association For Computational Linguistics}}.
\newblock


\bibitem[Wang et~al\mbox{.}(2024c)]%
        {wang-chartxiv-2024}
\bibfield{author}{\bibinfo{person}{Zirui Wang}, \bibinfo{person}{Mengzhou Xia}, \bibinfo{person}{Luxi He}, \bibinfo{person}{Howard Chen}, \bibinfo{person}{Yitao Liu}, \bibinfo{person}{Richard Zhu}, \bibinfo{person}{Kaiqu Liang}, \bibinfo{person}{Xindi Wu}, \bibinfo{person}{Haotian Liu}, \bibinfo{person}{Sadhika Malladi}, \bibinfo{person}{Alexis Chevalier}, \bibinfo{person}{Sanjeev Arora}, {and} \bibinfo{person}{Danqi Chen}.} \bibinfo{year}{2024}\natexlab{c}.
\newblock \showarticletitle{CharXiv: Charting Gaps in Realistic Chart Understanding in Multimodal LLMs}.
\newblock \bibinfo{journal}{\emph{CoRR}}  \bibinfo{volume}{abs/2406.18521} (\bibinfo{year}{2024}).
\newblock
\href{https://doi.org/10.48550/ARXIV.2406.18521}{doi:\nolinkurl{10.48550/ARXIV.2406.18521}}
\showeprint[arXiv]{2406.18521}


\bibitem[Wei et~al\mbox{.}(2022)]%
        {wei2022chain}
\bibfield{author}{\bibinfo{person}{Jason Wei}, \bibinfo{person}{Xuezhi Wang}, \bibinfo{person}{Dale Schuurmans}, \bibinfo{person}{Maarten Bosma}, \bibinfo{person}{Fei Xia}, \bibinfo{person}{Ed Chi}, \bibinfo{person}{Quoc~V Le}, \bibinfo{person}{Denny Zhou}, {et~al\mbox{.}}} \bibinfo{year}{2022}\natexlab{}.
\newblock \showarticletitle{Chain-of-thought prompting elicits reasoning in large language models}.
\newblock \bibinfo{journal}{\emph{Advances in neural information processing systems}}  \bibinfo{volume}{35} (\bibinfo{year}{2022}), \bibinfo{pages}{24824--24837}.
\newblock


\bibitem[Weng et~al\mbox{.}(2023)]%
        {weng2023large}
\bibfield{author}{\bibinfo{person}{Yixuan Weng}, \bibinfo{person}{Minjun Zhu}, \bibinfo{person}{Fei Xia}, \bibinfo{person}{Bin Li}, \bibinfo{person}{Shizhu He}, \bibinfo{person}{Shengping Liu}, \bibinfo{person}{Bin Sun}, \bibinfo{person}{Kang Liu}, {and} \bibinfo{person}{Jun Zhao}.} \bibinfo{year}{2023}\natexlab{}.
\newblock \showarticletitle{Large Language Models are Better Reasoners with Self-Verification}. In \bibinfo{booktitle}{\emph{Findings of the Association for Computational Linguistics: EMNLP 2023}}. \bibinfo{pages}{2550--2575}.
\newblock


\bibitem[Wolf et~al\mbox{.}(2019)]%
        {wolf2019huggingface}
\bibfield{author}{\bibinfo{person}{Thomas Wolf}, \bibinfo{person}{Lysandre Debut}, \bibinfo{person}{Victor Sanh}, \bibinfo{person}{Julien Chaumond}, \bibinfo{person}{Clement Delangue}, \bibinfo{person}{Anthony Moi}, \bibinfo{person}{Pierric Cistac}, \bibinfo{person}{Tim Rault}, \bibinfo{person}{R{\'e}mi Louf}, \bibinfo{person}{Morgan Funtowicz}, {et~al\mbox{.}}} \bibinfo{year}{2019}\natexlab{}.
\newblock \showarticletitle{Huggingface's transformers: State-of-the-art natural language processing}.
\newblock \bibinfo{journal}{\emph{arXiv preprint arXiv:1910.03771}} (\bibinfo{year}{2019}).
\newblock


\bibitem[Xia et~al\mbox{.}(2024)]%
        {xia2024chartx}
\bibfield{author}{\bibinfo{person}{Renqiu Xia}, \bibinfo{person}{Bo Zhang}, \bibinfo{person}{Hancheng Ye}, \bibinfo{person}{Xiangchao Yan}, \bibinfo{person}{Qi Liu}, \bibinfo{person}{Hongbin Zhou}, \bibinfo{person}{Zijun Chen}, \bibinfo{person}{Min Dou}, \bibinfo{person}{Botian Shi}, \bibinfo{person}{Junchi Yan}, {et~al\mbox{.}}} \bibinfo{year}{2024}\natexlab{}.
\newblock \showarticletitle{Chartx \& chartvlm: A versatile benchmark and foundation model for complicated chart reasoning}.
\newblock \bibinfo{journal}{\emph{arXiv preprint arXiv:2402.12185}} (\bibinfo{year}{2024}).
\newblock


\bibitem[Xu et~al\mbox{.}(2024)]%
        {xu2024llavacot}
\bibfield{author}{\bibinfo{person}{Guowei Xu}, \bibinfo{person}{Peng Jin}, \bibinfo{person}{Hao Li}, \bibinfo{person}{Yibing Song}, \bibinfo{person}{Lichao Sun}, {and} \bibinfo{person}{Li Yuan}.} \bibinfo{year}{2024}\natexlab{}.
\newblock \bibinfo{title}{LLaVA-CoT: Let Vision Language Models Reason Step-by-Step}.
\newblock
\showeprint[arxiv]{2411.10440}~[cs.CV]
\urldef\tempurl%
\url{https://arxiv.org/abs/2411.10440}
\showURL{%
\tempurl}


\bibitem[Xu et~al\mbox{.}(2023)]%
        {xu2023chartbench}
\bibfield{author}{\bibinfo{person}{Zhengzhuo Xu}, \bibinfo{person}{Sinan Du}, \bibinfo{person}{Yiyan Qi}, \bibinfo{person}{Chengjin Xu}, \bibinfo{person}{Chun Yuan}, {and} \bibinfo{person}{Jian Guo}.} \bibinfo{year}{2023}\natexlab{}.
\newblock \showarticletitle{Chartbench: A benchmark for complex visual reasoning in charts}.
\newblock \bibinfo{journal}{\emph{arXiv preprint arXiv:2312.15915}} (\bibinfo{year}{2023}).
\newblock


\bibitem[Yuan et~al\mbox{.}(2024)]%
        {yuan2024selfrewarding}
\bibfield{author}{\bibinfo{person}{Weizhe Yuan}, \bibinfo{person}{Richard~Yuanzhe Pang}, \bibinfo{person}{Kyunghyun Cho}, \bibinfo{person}{Xian Li}, \bibinfo{person}{Sainbayar Sukhbaatar}, \bibinfo{person}{Jing Xu}, {and} \bibinfo{person}{Jason~E Weston}.} \bibinfo{year}{2024}\natexlab{}.
\newblock \showarticletitle{Self-Rewarding Language Models}. In \bibinfo{booktitle}{\emph{Forty-first International Conference on Machine Learning}}.
\newblock
\urldef\tempurl%
\url{https://openreview.net/forum?id=0NphYCmgua}
\showURL{%
\tempurl}


\bibitem[Zhai et~al\mbox{.}(2024)]%
        {zhaifine2024}
\bibfield{author}{\bibinfo{person}{Simon Zhai}, \bibinfo{person}{Hao Bai}, \bibinfo{person}{Zipeng Lin}, \bibinfo{person}{Jiayi Pan}, \bibinfo{person}{Peter Tong}, \bibinfo{person}{Yifei Zhou}, \bibinfo{person}{Alane Suhr}, \bibinfo{person}{Saining Xie}, \bibinfo{person}{Yann LeCun}, \bibinfo{person}{Yi Ma}, {and} \bibinfo{person}{Sergey Levine}.} \bibinfo{year}{2024}\natexlab{}.
\newblock \showarticletitle{Fine-Tuning Large Vision-Language Models as Decision-Making Agents via Reinforcement Learning}. In \bibinfo{booktitle}{\emph{Advances in Neural Information Processing Systems}}, \bibfield{editor}{\bibinfo{person}{A.~Globerson}, \bibinfo{person}{L.~Mackey}, \bibinfo{person}{D.~Belgrave}, \bibinfo{person}{A.~Fan}, \bibinfo{person}{U.~Paquet}, \bibinfo{person}{J.~Tomczak}, {and} \bibinfo{person}{C.~Zhang}} (Eds.), Vol.~\bibinfo{volume}{37}. \bibinfo{publisher}{Curran Associates, Inc.}, \bibinfo{address}{Vancouver, Canada}, \bibinfo{pages}{110935--110971}.
\newblock
\urldef\tempurl%
\url{https://proceedings.neurips.cc/paper_files/paper/2024/file/c848b7d3adc08fcd0bf1df3101ba6728-Paper-Conference.pdf}
\showURL{%
\tempurl}


\bibitem[Zhang et~al\mbox{.}(2024)]%
        {zhang-etal-2024-llm-assisted}
\bibfield{author}{\bibinfo{person}{Meishan Zhang}, \bibinfo{person}{Gongyao Jiang}, \bibinfo{person}{Shuang Liu}, \bibinfo{person}{Jing Chen}, {and} \bibinfo{person}{Min Zhang}.} \bibinfo{year}{2024}\natexlab{}.
\newblock \showarticletitle{{LLM}-Assisted Data Augmentation for {C}hinese Dialogue-Level Dependency Parsing}.
\newblock \bibinfo{journal}{\emph{Computational Linguistics}} \bibinfo{volume}{50}, \bibinfo{number}{3} (\bibinfo{date}{Sept.} \bibinfo{year}{2024}), \bibinfo{pages}{867--891}.
\newblock
\href{https://doi.org/10.1162/coli_a_00515}{doi:\nolinkurl{10.1162/coli_a_00515}}


\bibitem[Zhao et~al\mbox{.}(2025)]%
        {zhao2025chartcoder}
\bibfield{author}{\bibinfo{person}{Xuanle Zhao}, \bibinfo{person}{Xianzhen Luo}, \bibinfo{person}{Qi Shi}, \bibinfo{person}{Chi Chen}, \bibinfo{person}{Shuo Wang}, \bibinfo{person}{Wanxiang Che}, \bibinfo{person}{Zhiyuan Liu}, {and} \bibinfo{person}{Maosong Sun}.} \bibinfo{year}{2025}\natexlab{}.
\newblock \bibinfo{title}{ChartCoder: Advancing Multimodal Large Language Model for Chart-to-Code Generation}.
\newblock


\end{thebibliography}

\end{document}